\documentclass[review,onefignum,onetabnum]{siamart171218}

\usepackage{lipsum}
\usepackage{amsfonts}
\usepackage{graphicx}
\usepackage{epstopdf}
\usepackage{algorithmic}
\usepackage{booktabs}
\usepackage{hyperref}
\ifpdf
  \DeclareGraphicsExtensions{.eps,.pdf,.png,.jpg}
\else
  \DeclareGraphicsExtensions{.eps}
\fi

\newsiamremark{remark}{Remark}
\newsiamremark{hypothesis}{Hypothesis}
\crefname{hypothesis}{Hypothesis}{Hypotheses}
\newsiamthm{claim}{Claim}

\headers{A NRDE-based model for path-dependent PDEs}{Bowen Fang, Hao Ni, and Yue Wu}

\title{A Neural RDE-based model for solving path-dependent parabolic PDEs\thanks{Submitted to the editors DATE.
\funding{HN is supported by the Engineering and Physical Sciences Research Council (EPSRC) [grant number EP/S026347/1] and the Alan Turing Institute under the EPSRC grant [grant number EP/N510129/1].}}}

\author{Bowen Fang\thanks{Department of Statistics, University of Warwick, Coventry, UK
  (\email{u2083124@live.warwick.ac.uk}).}
  \and Hao Ni\thanks{The Department of Mathematics, University College London, London, UK
  (\email{h.ni@ucl.ac.uk}).}
\and Yue Wu\thanks{Department of Mathematics and Statistics, University of Strathclyde, Glasgow, UK 
  (\email{yue.wu@strathclyde.ac.uk}).}}

\usepackage{amsopn}

\makeatletter
\newcommand*{\addFileDependency}[1]{
  \typeout{(#1)}
  \@addtofilelist{#1}
  \IfFileExists{#1}{}{\typeout{No file #1.}}
}
\makeatother

\newcommand*{\myexternaldocument}[1]{%
    \externaldocument{#1}%
    \addFileDependency{#1.tex}%
    \addFileDependency{#1.aux}%
}

\ifpdf
\hypersetup{
  pdftitle={A Neural RDE-based model for solving path-dependent PDEs},
  pdfauthor={Bowen Fang, Hao Ni and Yue Wu}
}
\fi
\usepackage{verbatim}

\myexternaldocument{ex_supplement}

\begin{document}
\nolinenumbers
\maketitle

\let\clearpage\relax

\begin{abstract}
The concept of the path-dependent partial differential equation (PPDE) was first introduced in the context of path-dependent derivatives in financial markets. Its semilinear form was later identified as a non-Markovian backward stochastic differential equation (BSDE).
Compared to the classical PDE, the solution of a PPDE involves an infinite-dimensional spatial variable, making it challenging to approximate, if not impossible. In this paper, we propose a  neural rough differential equation (NRDE)-based model to learn PPDEs, which effectively encodes the path information through the log-signature feature while capturing the fundamental dynamics. The proposed continuous-time model for the PPDE solution offers the benefits of efficient memory usage and the ability to scale with dimensionality. Several numerical experiments, provided to validate the performance of the proposed model in comparison to the strong baseline in the literature, are used to demonstrate its effectiveness. 
\end{abstract}



\section{Introduction}
The concept of path-dependent PDEs (PPDEs) arises from financial mathematics, when it comes to price path-dependent options such as Asian options and barrier options. Upon examining its solution via the functional Feynman-Kac formula developed in \cite{dupire_functional_2009} (see also Cont and Fournie \cite{cont_functional_2010,cont2010change, cont2013functional}), the PPDE is identical to the discounted price of an exotic option.  A more general PPDE can be formulated in the context of the nonlinear-Markovian backward stochastic differential equation (BSDE) through the same technique \cite{peng2010backward, peng2016bsde}. We also refer to \cite{ekren2014viscosity,ekren2016viscosity,ekren2016viscosity2,lukoyanov2007viscosity,peng2015g,cosso2015strong} for different approaches.

Solving PPDEs is practical but challenging, especially in high-dimensional cases. 
The capability of deep neural networks to approximate the function with complex structures, especially in high dimensional settings, could potentially tackle the "curse of dimensionality" faced by traditional numerical methods. Recent years have seen the development of numerous deep PDE solvers based on probabilistic reformulation \cite{Han_2018} \cite{https://doi.org/10.48550/arxiv.1804.07010}. Mathematical analysis for neural network approximations for PDEs could be found in \cite{E_2021}. Extending deep PDE solvers to PPDEs is however non-trivial, as 
the information of the path needs to be correctly encoded to the network, suggested by the name of PPDE. Some of the existing literature \cite{PDGM} utilizes long-short
term memory (LSTM) network \cite{hochreiter1997long} to capture the long-range dependence of the input data. 
Other work \cite{https://doi.org/10.48550/arxiv.2011.10630} relies on the path signature feature offered from rough path theory \cite{lyons_differential_2007} for capturing and condensing path characteristics, as 
it has been proven as a natural and effective feature in the machine learning context, from pattern recognition tasks \cite{handwriting_signature,Signature_humanaction} to healthcare applications \cite{wu_identifying_2022,Alzheimer,morrill_utilisation_nodate}. 
But the memory expense of both networks may not be scalable with dimensionality. 


To construct a suitable network structure that is memory-efficient and well-suited for the time series input, we exploit the neural rough differential equation (NRDE) network \cite{https://doi.org/10.48550/arxiv.2005.08926}, which is rooted in both rough path theory and neural ordinary differential equation (NODE), and propose an effective and hybrid model for solving high-dimensional PPDEs.  Consider a dynamical system that is governed by a controlled differential equation (CDE) with a control signal $X$ \cite{lyons_differential_2007}. 
Through a Lie algebra homomorphism on the truncated log-signature of the driving path $X$, one can replace the CDE with an autonomous ordinary differential equation (ODE) whose solution is a high-order approximation to the original solution of CDE \cite{gaines_ordinary_nodate}. Such a method is called the \emph{Log-ODE} method. Inspired by the NODE network introduced in \cite{chen2018neuralode}, 
Morrill et al. in \cite{NRDE} proposed a NRDE network which incorporates the autonomous ODE from the Log-ODE method, where part of the vector field is specified by a neural network. On the one hand, this network inherits most of the nice properties of the NODE network, such as memory efficiency through the use of the adjoint method (details can be found in Appendix \ref{sec:adjoint}) for calculating the gradient and adaptive computation. On the other hand, it extracts log-signatures of the input time series and has the advantage of tackling long time series. 


By interpreting the solution of the PPDE as a process of some evolving system governed by the path on which the PPDE depends, we build an NRDE-based hybrid model. This is the first time that NRDE is used to solve PPDEs.
The main features of our proposed model include:
\begin{itemize}
\item The model captures the dynamics of the PPDE and preserves intrinsic features of the underlying path;
\item Unlike existing literature, the log-signature is included as a layer rather than exacted features;
\item It is capable of solving both low-dimensional and high-dimensional PPDEs, and, compared to the baseline model, the benefit becomes more significant as dimension increases;
    \item Its computational efficiency can be further enhanced by introducing an embedding layer;
    \item  It admits flexible time steps in solving the ODE of the NRDE network, which enables easy adjustment of the model complexity based on task needs;
    \item It possesses the characteristic of the NRDE, which entails a constant memory expense.
\end{itemize}

The problem setup is as follows: given $T>0$ and $d,m\in \mathbb{N}$,  we are interested in simulating 
 $u:[0, T] \times C\!\left([0, T], \mathbb{R}^{d}\right) \rightarrow \mathbb{R}$,\footnote{$C\!\left([0, T], \mathbb{R}^{d}\right)$ is the collection of continuous functions mapping from $[0,T]$ to $\mathbb{R}^d$.} which is the solution of the following semilinear PPDE with a terminal condition,
\begin{align}\label{eqn:ppde}
\begin{split}
     &\left[\partial_{t} u+b D_x u+\frac{1}{2} \operatorname{tr}\left[D_{xx} u \sigma^{T} \sigma\right]-r u\right](t, \omega)+f(t,\omega)=0\\
    &u(T, \omega)=g(\omega), \quad t \in[0, T], \quad \omega \in C\!\left([0, T] ; \mathbb{R}^{d}\right),
\end{split}
\end{align}
where $b$, $\sigma$, $r$ and $f$ are functions that map from $[0, T] \times C\!\left([0, T], \mathbb{R}^{d}\right)$ to $\mathbb{R}^d,\mathbb{R}^{d\times m},\mathbb{R}$ and $\mathbb{R}$ respectively, and $D_x u$ and $D_{xx}u$ are path derivatives that will be specified later. 

The remaining paper is structured as follows: in the Preliminaries section (Section \ref{sec:pre}), a concise overview of the fundamental notations in functional It\^o calculus is presented, with specific emphasis on the Feynman-Kac formula, which is followed by an short introduction to (log-)signatures. Section \ref{sec:algo} reviews two learning frameworks that we will adopt. The NRDE-based hybrid model architecture for solving high-dimensional PPDEs is proposed in Section \ref{sec:network} with numerical experiments in Section \ref{sec:exp}. Across all experiments, our model consistently surpasses the performance of the baseline model, demonstrating its remarkable capabilities. Discussion, limitation and future works are included is given in Section \ref{sec:conc}. 


\section{Preliminaries} \label{sec:pre}
This section is devoted to a brief summary of the basics of the functional It\^o calculus and rough path theory. We begin with definitions of both the time and spacial derivative of a path-dependent functional and then the functional It\^o formula. This is followed by the probabilistic representation of the solution of the PPDE \eqref{eqn:ppde} via the functional Feynman-Kac formula. One may refer to \cite{dupire_functional_2009} for further details of the functional It\^o calculus. In Section \ref{subsec:rough path}, we introduce the path signature and log-signature features, which are the core concepts of rough path theory and serve as effective feature representations of sequential data. 

\subsection{Functional It\^o formula}
We start by defining the path space. Fix $T>0$ to be the maturity time. For every $t\in [0,T]$, denote $\Lambda_{t}$ the set of \text{càdlàg}\footnote{It means the function is right continuous with left limits.} $\mathbb{R}^d$-valued functions on $[0,t]$. For $\omega \in \Lambda_{T}$, the value of $\omega$ at time $t\in [0,T]$ is denoted by $\omega(t)$. To distinguish $\omega(t)$, we define the path $\omega$ restricted to the time interval $[0, t]$ by $\omega_{[0,t]}$. The path space, denoted by $\Lambda$, consists of all possible paths $\omega \in \Lambda_{t}$ for $t \in [0, T]$; in formula, $\Lambda=\bigcup_{t \in[0, T]} \Lambda_{t}$. Let $\langle\cdot,\cdot\rangle$ and $|\cdot|$ be the inner product and the norm in $\mathbb{R}^d$. The norm on the path space $\Lambda$ inherits the supermum norm of the continuous space, i.e.,
$$\left\|\omega_{[0,t]}\right\|:=\sup _{r \in[0, t]}|\omega(r)|.$$ 
For $0\leq t\leq \hat{t}\leq T$ and $\omega_{[0,t]},\hat{\omega}_{[0,\hat{t}]}\in \Lambda$, we can also characterise their distance in $\Lambda$ by
\begin{align*}
    &d_{\infty}\!\left(\omega_{[0,t]}, \hat{\omega}_{[0,\hat{t}]}\right):=\max \left(\sup _{r \in[0, t)}|\omega(r)-\hat{\omega}(r)|, \sup _{r \in[t, \hat{t}]}|\omega(t)-\hat{\omega}(r)|\right)+|t-\hat{t}|.
\end{align*}

Let $u: \Lambda \rightarrow \mathbb{R}$ denote a functional mapping any \text{càdlàg} path over $[0,t]$ to a real number, where $t\in [0, T]$. 
The functional $u: \Lambda \mapsto \mathbb{R}$ is said to be $\Lambda$-continuous at $\omega_{[0,t]}\in \Lambda$ if for any $\varepsilon>0$, there exists $\delta>0$ such that $\forall \hat{\omega}_{[0,\hat{t}]}\in\Lambda$ satisfying $d_{\infty}\big(\omega_{[0,t]},\hat{\omega}_{[0,\hat{t}]}\big)<\delta$, we have $|u(\omega_{[0,t]})-u(\hat{\omega}_{[0,\hat{t}]})|<\epsilon$. We say that $u$ is $\Lambda$-continuous if it is continuous at all $\omega_{[0,t]}\in \Lambda$.
    
\begin{figure}
    \centering
    \includegraphics[scale=0.15]{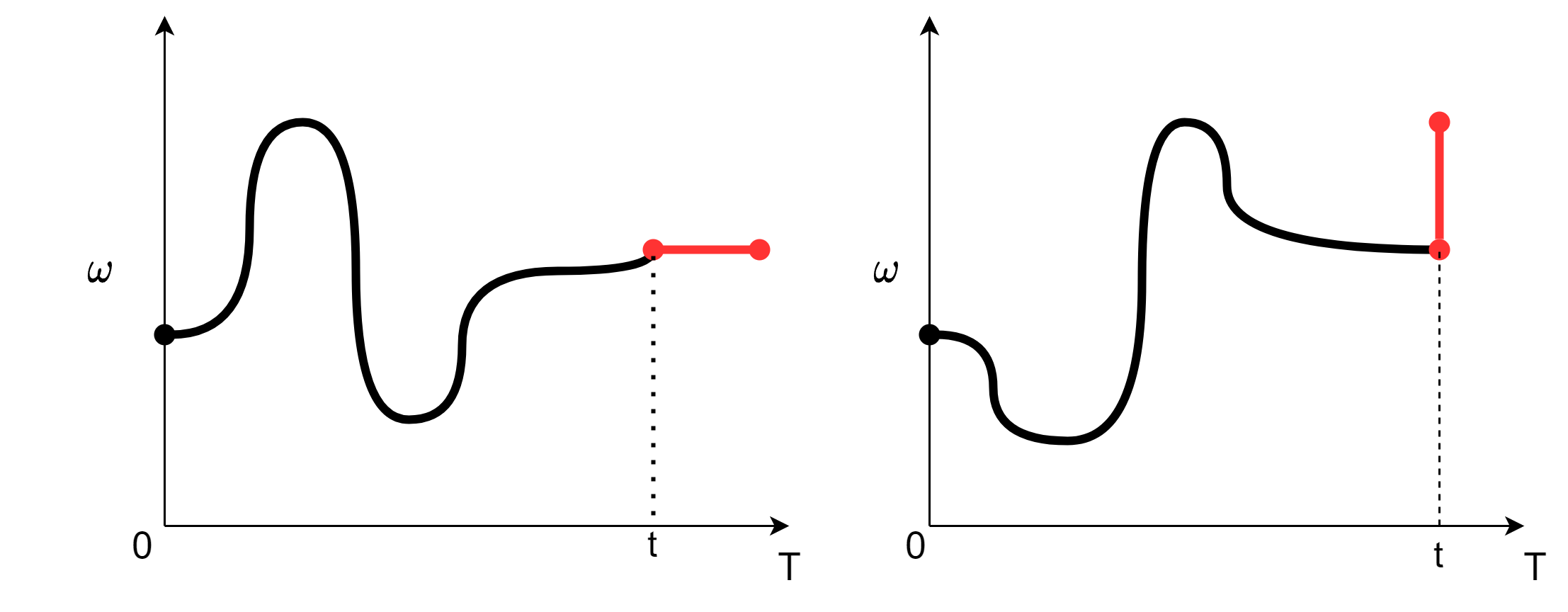}
    \caption{Illustration of the flat extension $\omega_{[0,t],s}$ (left) and jump extension $\omega_{[0,t]}^{x}$(right) of $\omega_{[0,t]}$.\label{fig:path_ext} }
\end{figure}

We now define the space and time derivatives of functional $u: \Lambda \rightarrow \mathbb{R}$. For a path $\omega_{[0,t]} \in \Lambda$ and $x\in \mathbb{R}^d$, we denote the jump extension 
\begin{align*}
    w^{x}_{[0,t]}(s)=\Big\{\begin{array}{l} w(s),\;\;\; 0\leq s<t;\\ w(t)+x \;\;\; s=t.  \end{array}
\end{align*}
The right subplot of Fig \ref{fig:path_ext} gives an example of a jump extension for a one-dimensional path $\omega_{[0,t]}$. Now let $u: \Lambda \mapsto \mathbb{R}$ and $\omega_{[0,t]}\in \Lambda$, the space derivative $D_{x} u\!\left(\omega_{[0,t]}\right)$ at $\omega_{[0,t]}$ is $p\in \mathbb{R}^d$ such that 
\begin{equation}   u\!\left(\omega_{[0,t]}^{x}\right)=u\!\left(\omega_{[0,t]}\right)+\langle p, x\rangle+o(|x|), x \in \mathbb{R}^{d}.
\end{equation}
We say that $u$ is \emph{vertically differentiable} in $\Lambda$ if $D_{x} u\!\left(\omega_{[0,t]}\right)$ exists for every $\omega_{[0,t]}\in \Lambda$. The second order derivative $D_{x x} u\!\left(\omega_{[0,t]}\right)$ is a $d \times d$ symmetric matrix defined similarly.

Let $\omega_{[0,t],s}$ with $0\leq t<s\leq T$ represent an extension of $\omega_{[0,t]}$ such that the value at $\omega(t)$ is frozen over $[t,s]$. The left subplot of Fig \ref{fig:path_ext} gives an example of such an extension for a one-dimensional path $\omega_{[0,t]}$. For a given $\omega_{[0.t]}\in \Lambda$, we say that $u(\omega_{[0,t]})$ is \emph{horizontally differentiable} in $t$ at $\omega_{[0,t]}$ if there exists $D_{t} u\!\left(\omega_{[0,t]}\right)=a$ such that $u\!\left(\omega_{[0,t], s}\right)=u\left(\omega_{[0,t]}\right)+a(s-t)+o(|s-t|)$. 
The functional $u$ is said to be horizontally differentiable in $\Lambda$ if $D_{t} u\left(\omega_{[0,t]}\right)$ is well defined for all $\omega_{[0,t]} \in \Lambda$.\\
Define by $\mathbb{C}^{j, k}(\Lambda)$ the set of functionals $u$ on $\Lambda$, which are $j$-times horizontally and $k$-times vertically differentiable in $\Omega$ such that all these derivatives are $\Lambda$-continuous. Further, define by $\mathbb{C}^{i, j, k}(\Lambda)$ the set of $u: [0,T]\times\Lambda \to \mathbb{R}$, where $u$ is also $i$-times differentiable in time. For $s>t$, we denote by $b:=\partial_t u\!\left(t,\omega_{[0,t]}\right)$ the first derivative of $u$ with respect to time such that
$$u\!\left(s,\omega_{[0,t],s}\right)=u\!\left(t,\omega_{[0,t]}\right)+b(s-t)+o(|s-t|).$$
Note that here $b$ has included the horizontal derivative defined before.

We summarize the functional It\^o formula \cite{dupire_functional_2009, cont_functional_2010} as follows.
\begin{theorem}(Functional It\^o formula). Let $\left(\Omega, \mathcal{F}, \mathbb{P}\right)$ be a probability space and $(\mathcal{F}_t)_{t\in [0,T]}$ be the filtration defined on this probability space. If $X$ is a continuous semi-martingale and $u$ is in $\mathbb{C}^{1,1,2}(\Lambda)$, then for any $t\in [0,T)$, the following Eqn. \eqref{eqn:functional-Ito} holds almost surely,
\begin{align}\label{eqn:functional-Ito}
   \begin{split}
          u\!\left(t,X_{[0,t]}\right)-u\!\left(0,X_{0}\right)&=\int_{0}^{t} \partial_{s} u\!\left(s,X_{[0,s]}\right)\, \mathrm{d} s+\int_{0}^{t} D_{x} u\left(s,X_{[0,s]}\right) \mathrm{d} X(s) \\
    &\quad
           +\frac{1}{2} \int_{0}^{t} D_{x x} u\left(s,X_{[0,s]}\right) \mathrm{d}\langle X\rangle(s),
   \end{split}
\end{align}
where $\langle X \rangle(t)$ is the quadratic variation process of $X$, defined as 
\begin{equation*}
    \langle X\rangle(t)=\lim_{|\Pi_n|\to 0}\sum_{k=1}^{n}(X(t_{k})-X(t_{k-1}))(X(t_{k})-X(t_{k-1}))^T,
\end{equation*}
where the limit is taken in the $L^2(\Omega)$ sense and taken over all possible time partitions\footnote{ We say that $\Pi_n$ is a partition of $[0, t]$ if $\Pi_n=\{(t_i)_{i = 1}^{n} | 0=t_0 \leq \dots \leq t_n=t\}$. $|\Pi_n|$ represents the largest mesh size of the partition $\Pi_n$ } of $[0,t]$.
\end{theorem}
\subsection{Feynman-Kac formula for functionals} Given a complete probability space $(\Omega,\mathcal{F},\mathbb{P})$ and a
standard Wiener process $W \colon [0,T] \times \Omega \to \mathbb{R}^m$ on $(\Omega, \mathcal{F}, \mathbb{P})$. For a random path $X:[0, T] \times \Omega \rightarrow \mathbb{R}^{d}$, denote by $X_{[0,t]}$ the stopped process of $X$ at time $t$. Let $\sigma$ and $b$ be the functionals of the pair $(t, X_{[0,t]})$ such that the following stochastic functional differential equation (SFDE) is well defined:
\begin{equation}\label{eqn:sfde}\mathrm{d}X(t)=b(t,X_{[0,t]})\,\mathrm{d}t+\sigma(t,X_{[0,t]})\,\mathrm{d}W(t).
\end{equation}
Then the solution of equation \eqref{eqn:ppde} has a probabilistic representation via the solution of an SFDE \cite{dupire_functional_2009,book}.
\begin{theorem}[Feynman-Kac formula]\label{thm:feyman}
If $u$ is a non-anticipative functional, i.e., for any given $t\in [0, T]$ it holds that $u(t,\omega^1_{[0,T]})=u(t,\omega^2_{[0,T]})$ if $\omega^1_{[0,t]}=\omega^2_{[0,t]}$; then given any solution $u(t,\omega) \in \mathbb{C}^{1,1,2}$ that satisfies
\begin{align}
\begin{split}
     &\left[\partial_{t} u+b D_x u+\frac{1}{2} \operatorname{tr}\left[D_{xx} u \sigma^{T} \sigma\right]-r u\right](t, \omega)+f(t,\omega)=0\\
    &u(T, \omega)=g(\omega), \quad t \in[0, T], \quad \omega \in C\!\left([0, T] ; \mathbb{R}^{d}\right),
\end{split}
\end{align}
it has the representation
\begin{align}\label{eqn:feynman}
       u(t, \omega)=&\mathbb{E}\Big[g(X_{[0,T]})e^{-\int_t^T r(s,X_{[0,s]})\,\mathrm{d}s}\nonumber \\
       &+\int_t^T f\!\left(s,X_{[0,s]}\right)e^{-\int_t^s r(h,X_{[0,h]})\,\mathrm{d}h}\,\mathrm{d}s \big|X_{[0,t]}=\omega_{[0,t]}\Big],
\end{align}
where $X$ satisfies the SFDE \eqref{eqn:sfde}.
\end{theorem}
\subsection{Path signature and log-signature for time series data}\label{subsec:rough path} This section presents the introduction of path signature and log-signature features, providing a summary of their key properties and a comparison between them.

Let us first introduce the tensor algebra space, where the signature of a path takes value. Denote $E$ the Banach space in which the paths take their values, we could assume it has a finite-dimensional structure. In our case, $E=\mathbb{R}^d$ or $\mathbb{R}^m$ with the canonical Euclidean norm. The corresponding tensor algebra space is defined to be $T((E)):=\oplus_{n=0}^\infty E^{\otimes n}$. For details please refer to Appendix \ref{sec:signature}.

\begin{definition}[Signature of a path]
Let $J$ denote a compact time interval and $X:J\to E$ be a continuous path. The signature of $X$ is an element of $T((E))$ defined as
\begin{equation*}
    S_J(X)=\left(1, X_J^1, X_J^2, \ldots\right),
\end{equation*}
where, for each $n\geq 1$, 
\begin{equation}\label{eqn:ite_int}
    X_J^n:=\int_{\substack{u_1<\ldots<u_n \\ u_1, \ldots, u_n \in J}} \mathrm{d} X_{u_1} \otimes \ldots \otimes \mathrm{d} X_{u_n} \in E^{\otimes n}.
\end{equation}
The truncated signature of $X$ of depth $N$ is defined as $$S^{N}_J(X):=(1,X^1_J,\dots,X^N_J).$$
\end{definition}

The rough path theory shows that the solution of a controlled system driven by path $X$ is uniquely determined by its signature and the initial condition \cite{lyons_differential_2007}.  The signature of a path with finite length is a fundamental representation that captures its impact on any nonlinear system. It enables the local approximation of path or stream effects through linear combinations of signature elements. Consequently, the coordinate iterated integrals, or the signature as a whole, form a natural feature set for capturing the data aspects that predict the impact of the path on a controlled system.

On the other hand, two distinct paths can
have exactly the same signature. For example, they are the same under time reparameterization. 
However, Hambly and Lyons \cite{hambly_uniqueness_2010} show that the signature completely determines the path's geometry up to a tree-like equivalence.  In addition, a time-augmented path can be uniquely determined by and recovered from its signature \cite{levin2013learning}.

The log-signature is a parsimonious description of the signature, while the (truncated) log-signature and signature are bijective. To define the log-signature, we need to introduce the logarithm of an element in the tensor algebra space. For any $x=\left(1, x_1, \cdots, x_n,\cdots\right) \in T((E))$ , the logarithm of $x$ is defined as 
\begin{equation}\label{eqn_tensor_logarithm}
    \log (x)=\sum_{n=1}^{\infty} \frac{(-1)^{n-1}}{n} \left(x-1\right)^{\otimes n},
\end{equation}
where the multiplication involved is the tensor multiplication. 
\begin{definition}[Log-signature of a path]
Let $X: [0, T] \rightarrow \mathbb{R}^d$ be a continuous path and the signature of $X$ is well defined. Then the log-signature of path $X$, denoted as $\operatorname{Logsig}(X)$, is the logarithm of the signature of the path $X$.   
\end{definition}

Denote by $\operatorname{Logsig}^{N}(X)$ the truncated log-signature of depth $N$. The dimension of this truncated log signature, denoted by $\beta(d,N)$ depends on the dimensionality of the path $d$ and the truncation depth $N$ (see appendix \ref{sec:signature} for its formula). In contrast to the signature, the log-signature offers the benefit of dimension reduction thus reduces feature redundancy, but it should be combined with non-linear models for approximating any functional on the unparameterized path space \cite{liao2019learning}.
One can refer to \cite{DBLP:conf/bmvc/NiLYSL21,lyons_differential_2007} or the appendix \ref{sec:signature} for further details of the log-signature \cite{levin2013learning}. 

Several Python packages (i.e., ESig \cite{esig}, iisignature \cite{iisig}, Signatory \cite{signatory}) are available for calculating signature and log-signature features from discrete time series data.  

\section{Learning PPDE solution via a supervised learning approach}\label{sec:algo}
This section reviews the existing learning frameworks \cite{https://doi.org/10.48550/arxiv.2005.08926} that we will adopt to learn the solution of the target PPDE. The first framework directly builds the loss function on the Feynman-Kac representation of the solution, whereas the subsequent framework leverages an additional property of the spatial derivative in addition to the primary one, but is subject to certain conditions imposed on $f$ and $r$.
\subsection{Method 1: learning the solution as a conditional expectation}\label{sec:method1}
we define the entire term within the conditional expectation \eqref{eqn:feynman} as $F$:
\begin{equation}\label{eqn:Fdef}
    F(t,X):=g(X_{[0,T]})e^{-\int_t^T r(s,X_{[0,s]})\mathrm{d}s}+\int_t^T f\!\left(s,X_{[0,s]}\right)e^{-\int_t^s r(h,X_{[0,h]})\mathrm{d}h}\mathrm{d}s.
\end{equation}
Suppose $\left(\mathcal{F}_{t}\right), t \geq 0$ is a filtration generated by $\left(X_{t}\right)_{t \geq 0}$, and denote $L^2(\mathcal{F}_t)$ the space of square integrable and $\mathcal{F}_t$ measurable random variable. Then the conditional expectation \eqref{eqn:feynman} is the function $h \in L^{2}\!\left(\mathcal{F}_{t}\right)$ that gives the best approximate of $F(t,X)$ in the mean quadratic sense, that is
\begin{align}
    u(t, \omega)&= \mathbb{E}\Big[F(t,X)\Big|X_{[0,t]}=\omega_{[0,t]}\Big]=\inf _{h \in {L}^{2}\!\left(\mathcal{F}_{t}\right)}\mathbb{E}\Big[\Big|F(t,X)-h\Big|^2\Big]\label{eqn:cond_exp}.
\end{align}
By Doob-Dynkin lemma \cite{doob_stochastic_1953}, one can always express the function $h$ as $\hat{h}(t, X_{[0,t]})$ for some $\mathcal{F}_t$-measurable function $\hat{h}$. We will approximate such $\hat{h}$ by a neural network $\hat{u}_{\theta}$. Eqn. \eqref{eqn:cond_exp} thus provides a learning task for trainable parameters $\theta$ of neural network.

Now fix the terminal time $T$ and we are interested in the value of $u(t,X)$ on some time grid $0=t_0\leq t_1<t_2<\dots<t_{N_1}=T$ with $N_1\in \mathbb{N}$. We first generate $N_2$ with $N_2\in \mathbb{N}$ many trajectories of underlying $X$ from \eqref{eqn:sfde} via some numerical scheme, evaluate $F$ upon these realizations, then find the optimal $\hat{u}_{\theta}$ by minimizing the difference between $\hat{u}_{\theta}$ and $F$. The objective function for the trainable parameter $\theta$ is therefore 
\begin{equation}\label{eqn:theta*}
    \theta^{*}=\underset{\theta}{\arg \min }\frac{1}{N_2}\sum_{i=1}^{N_2}\sum_{j=0}^{N_1}\left(F\big(t_j,X^{(i)}\big)-\hat{u}_{\theta}\big(t_{j},X_{[0,t_j]}^{(i)}\big)\right)^{2},
\end{equation}
where $X^{\left(i\right)}$ represents the $i^{\text{th}}$ sampled path trajectory of $X$.
\subsection{Method 2: learning both the PPDE solution and path-dependent derivative}
Now consider the special case of Eqn. \eqref{eqn:ppde} when $f=0$ and $r$ is a constant. This can be related to a simple scenario in the context of financial mathematics, where the underlying assets $X$ admit the following SDE: 
\begin{equation}
    \mathrm{d}X(t)=rX(t)\mathrm{d}t+\sigma(t,X_{[0,t]})\mathrm{d}W(t).
\end{equation}
Note that the discounted price process $\bar{X}(t):=e^{-rt}X(t)$ now gives a martingale. Let $g(X_{[0,T]})$ be the final payoff of some derivative that depends on $X$, and define the square-integrable process
\begin{equation*}
    M(t):=\mathbb{E}\left[e^{-r T} g(X_{[0,T]}) \mid \mathcal{F}_{t}\right].
\end{equation*}
The fair price of the derivative at time $t$ is $u(t,X_{[0,t]}):=e^{rt}M(t)$. By the martingale representation theorem \cite{revuz_continuous_1999} and the functional It\^o formula \eqref{eqn:functional-Ito}, there exists a unique adapted process $Z$ such that
\begin{equation*}
    M(T)=\mathbb{E}\left[M(T) \mid \mathcal{F}_{0}\right]+\int_{0}^{T} Z(t)\mathrm{d} W(t)=\mathbb{E}\left[M(T) \mid \mathcal{F}_{0}\right]+\int_{0}^{T} D_x u(t,X_{[0,t]}) \mathrm{d} \bar{X}(t).
\end{equation*}
This leads to the following relation: for  $1\leq j\leq N_1$,
\begin{equation}
   M(t_{j})=M(t_{j-1})+\int_{t_{j-1}}^{t_{j}} D_{x} u(v,X_{[0,v]}) \,\mathrm{d} \bar{X}(v),
\end{equation}
and allows us to add another network  $\left[D_{x}\hat{u}\right] _{\phi}$ to approximate the path derivative separately.

If $\hat{u}_{\theta}$ and $\left[D_{x}\hat{u}\right] _{\phi}$ both provide good approximations, then at each $t_j$, the terms
$$e^{-rt_j}\hat{u}_{\theta}\big(t_j,X_{[0,t_j]}\big)-e^{-rt_{j-1}}\hat{u}_{\theta}\big(t_{j-1},X_{[0,t_{j-1}]}\big)-\int_{t_{j-1}}^{t_{j}} \left[D_{x}\hat{u}\right] _{\phi}(v,X_{[0,v]}) \,\mathrm{d} \bar{X}(v)$$
should be close to zero. Under such a consecutive relation, only the terminal condition need to taken into account in \eqref{eqn:theta*}. The goal now is to minimize two terms simultaneously 
\begin{align}\label{eqn:theta_method2}
     &(\theta^{*},\phi^{*})\nonumber\\
     &=\underset{\theta,\phi}{\arg \min }\frac{1}{N_2}\sum_{i=1}^{N_2}\sum_{j=0}^{N_1}\Big\{\big[e^{-rt_j}\hat{u}_{\theta}\big(t_j,X^{(i)}_{[0,t_j]}\big)-e^{-rt_{j-1}}\hat{u}_{\theta}\big(t_{j-1},X^{(i)}_{[0,t_{j-1}]}\big)\\
     &\hspace{3.4cm}-\left[D_{x}\hat{u}\right]_{\phi}\Delta \bar{X}^{(i)}_{j} \big]+\left(g(T,X^{(i)}_{[0,T]})-\hat{u}\big(T,X^{(i)}_{[0,T]}\big) \right)^2\Big\},\nonumber
\end{align}
where $\Delta \bar{X}_{j}=\bar{X}_j-\bar{X}_{j-1}$.
\section{Network architecture for the PPDE solution} \label{sec:network}
In this section, we propose a network architecture
to handle high dimensional PPDEs with a dimension-reduction feature. To be more precise, the approximation network $\hat{u}_{\theta}(t,X)$ is assumed to follow a controlled differential equation (CDE) driven by $X$ but with an unknown vector field, which can be solved through a neural rough differential equation (NRDE) network \cite{NRDE} with truncated log-signature features of $X$. When the driving path $X$ is of high dimension, an embedding layer is introduced for dimension reduction.  

\subsection{The NRDE network}
Given $t_0,T\in \mathbb{R}$ with $t_0<T$ and $d,h\in \mathbb{N}$. Let $\xi \in \mathbb{R}^{h}$, $X:[t_0,T]\rightarrow\mathbb{R}^d$ be a continuous function of bounded variation  and $G:\mathbb{R^d}\rightarrow \mathbb{R}^{h\times d}$ be a continuous function.Consider $Z:[t_0,t_1]\rightarrow \mathbb{R}^h$ which is the unique solution to the following CDE
\begin{equation}\label{eqn:cde}
    Z(t_0)=\xi,\;Z(t)=Z(t_0)+\int_{t_0}^t G(Z(s))\,\mathrm{d}X(s)\ \text{for } t\in (t_0,T],
\end{equation}
where $G(Z(s))\,\mathrm{d}X(s)$ is understood as a matrix-vector product. 

Now given a time series data $X=((t_0,x_0),(t_1,x_1),\dots,(t_n,x_n))$ and we are interested the output of the model at specific points $t_0=r_0<r_1<\dots<r_m=t_n\leq T$. 
To achieve this, inspired by the Log-ODE method \cite{gaines_ordinary_nodate}, the NRDE network gives the approximation iteratively by replacing the vector field $f(Z(s))\,\mathrm{d}X(s)$ in \eqref{eqn:cde} through the following piecewise function
\begin{equation}\label{eqn:g_nrde}
    \hat{g}_{\theta, X}(Z, s)=\hat{G}_{\theta}(Z) \frac{\operatorname{LogSig}_{r_{i}, r_{i+1}}^{N}(X)}{r_{i+1}-r_{i}} \text { for } s \in\left[r_{i}, r_{i+1}\right),
\end{equation}
where $\hat{G}_{\theta}:\mathbb{R}^h\rightarrow\mathbb{R}^{h\times \beta(d, N)}$ is an arbitrary neural network with trainable parameters $\theta$, $\operatorname{LogSig}_{r_{i}, r_{i+1}}^{N}(X) \in \mathbb{R}^{\beta(d,N)}$ is the depth-$N$ truncated log-signature of $X$ over the interval $[r_i,r_{i+1}]$, and the right-hand side of \eqref{eqn:g_nrde} is a matrix-vector product. In addition, a layer $\xi_{\theta}:\mathbb{R}^d\rightarrow \mathbb{R}^h$ is added to calculate the initial hidden state $Z_{t_0}$. The Eqn. \eqref{eqn:cde} thus boils down to an ordinary differential equation (ODE)
\begin{equation}\label{eqn:NRDE}
   Z(t_0)=\xi_{\theta}(X(t_0)),\;\; Z(t)=Z_{t_0}+\int_{t_{0}}^{t} \hat{g}_{\theta, X}\!\left(Z(s), s\right) \mathrm{d} s,
\end{equation}
where the universality is gaurenteed in \cite{https://doi.org/10.48550/arxiv.2005.08926} and \cite{NRDE}.

The implementation of \eqref{eqn:NRDE} can be done via existing ODE solver package such as \textit{torchdiffeq} \cite{chen2018neuralode,chen2021eventfn}. The training of parameters $\theta$ is done using stochastic gradient descent in the usual way, while we can carry out the forward pass and backpropagation via adjoint methods, which have the memory efficient advantage.
\subsection{Using embedding for dimension reduction}
\begin{figure}
    \centering
    \includegraphics[scale=0.1]{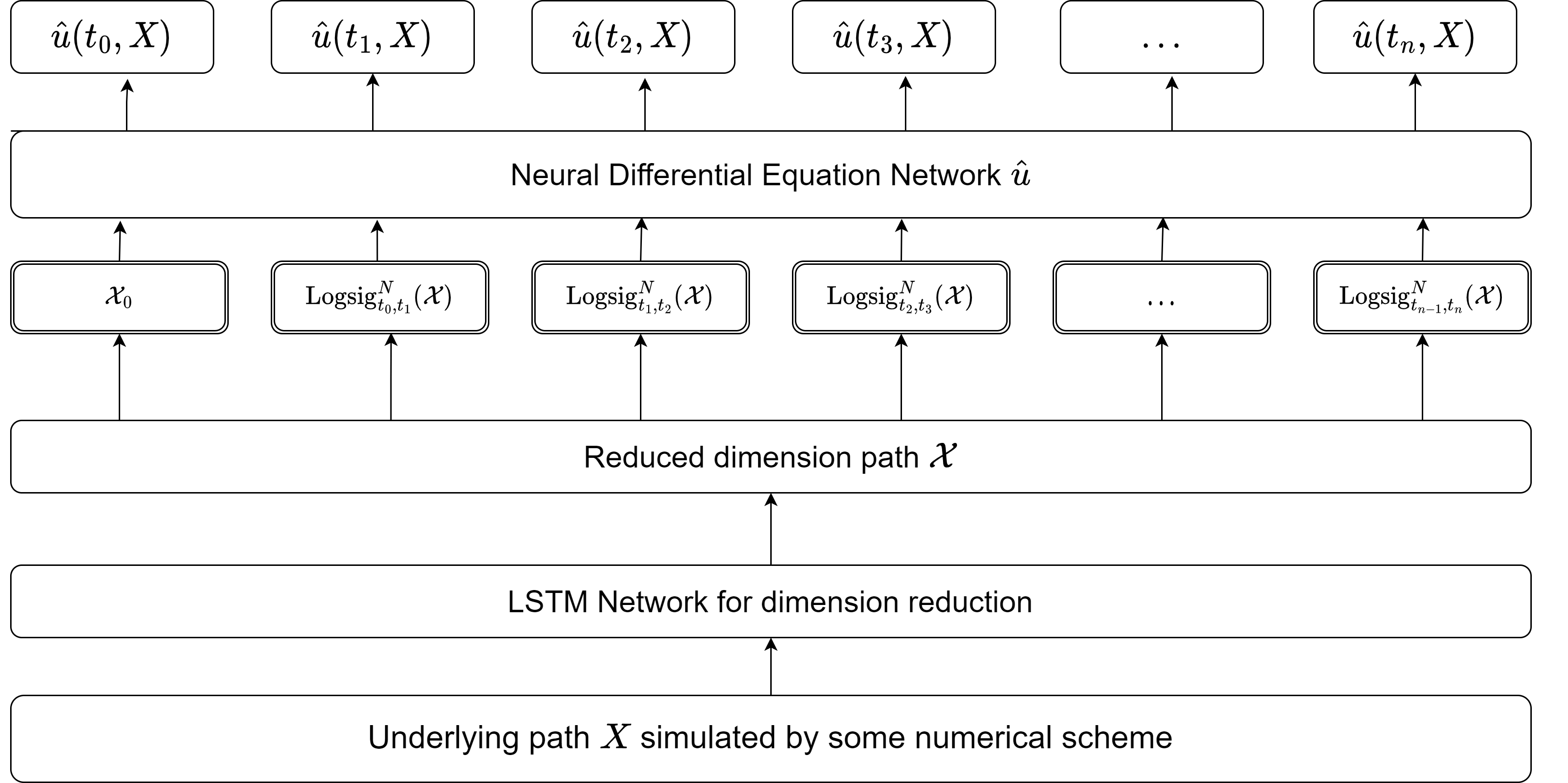}
    \caption{The Structure of EL-NRDE model.}
    \label{fig:LSTM-RNDE}
\end{figure}

In our model, the signature method is used to capture the dynamics of path $X:[0,T]\rightarrow \mathbb{R}^{d}$ between consecutive time steps, but the scalability of the model becomes a challenge when the dimension increases. In particular, for a fixed depth $N$, the size of $\operatorname{Sig}^{N}(X)$ and $\operatorname{Logsig}^{N}(X)$ increase exponentially with dimension\footnote{One may refer to \cite[Table 2]{reizenstein2017calculation} for the collection of size numbers of truncated (log-)signature of depth upto $12$ for paths of dimension between $1$ to $5$.}, leading to a heavy computational cost for the model. For instance, fix $N=3$, the sizes of the truncated signatures of depth $3$ are $14, 39, 84$ and $155$ respectively for paths of dimension $2$ to $5$, and the corresponding sizes of truncated log-signature of depth $3$ are $5, 14, 30$ and $55$.   To address this issue, we introduce an additional embedding layer to reduce the dimension of $X$ and transform $X\rightarrow \mathcal{X}$ when necessary. The resulting $\mathcal{X}:[0,T]\rightarrow \mathbb{R}^{d_1}$ has a smaller dimension $d_1<d$, and the truncated log-signature of $\mathcal{X}$ is calculated and fed into the subsequent trainable layers,  as illustrated in Figure \ref{fig:LSTM-RNDE}. The proposed model is referred to as \emph{EL-NRDE}. Note that the dimension reduction feature is also effective for the LSTM model in \cite{https://doi.org/10.48550/arxiv.2005.08926}, as will be shown in the experiment section.

\section{Experimental results} \label{sec:exp}
This section demonstrates the capability of the proposed NRDE model through diverse experiments of PPDEs. To showcase the scalability of the model, each experiment except the last one starts with a low-dimensional example and progresses to high-dimensional cases. We compare the performance of our model with respect to the LSTM model with signature features proposed in \cite{https://doi.org/10.48550/arxiv.2005.08926}, referred to as \emph{SigLSTM}. We have selected this model as our baseline for two main reasons. Firstly, this model has demonstrated state-of-the-art performance, indicating its effectiveness in solving PPDEs. Secondly, both our model and the baseline model make use of signature methods. This commonality allows for a fair and meaningful comparison between the two approaches. By leveraging signature methods, both models may share similarities in their underlying principles or techniques, which can help to compare their strengths, weaknesses, and overall effectiveness. By considering these two factors, we can establish a solid foundation for evaluating our model's performance and drawing meaningful insights and conclusions in comparison to the baseline model.

As aforementioned, for high-dimension PPDEs, it is necessary to incorporate an embedding layer prior to the model. We, therefore, prefix the model with the dimension reduction layer with the name 'EL'. For example,  \emph{EL-SigLSTM} refers to the model that combines the SigLSTM network and the dimension reduction feature.
\subsection{Experimental setup}
\subsubsection{Training procedure}
We set the training step to $2000$ epochs. During each epoch, some suitable numerical method such as Euler-Maruyama method is used to simulate a batch of trajectories of the underlying SDE, then adopt either \eqref{eqn:theta*} or \eqref{eqn:theta_method2} as the objective function to update the network parameters.
\emph{Adagrad} is applied as the optimizer during training, and the adjoint method developed in \cite{chen2018neuralode} is chosen to calculate the gradient with memory efficiency. If the vector field $\hat{f}$ requires $\mathcal{O}(H_1)$ memory, and the length of time series is $H_2$, then backpropagating through the solver requires $\mathcal{O}(H_1 H_2)$ memory while the adjoint method requires $\mathcal{O}(H_1+H_2)$. We include some details about the adjoint method in appendix \ref{sec:adjoint} and this has also been discussed in \cite{chen2018neuralode, kidger2021hey}. 
\subsubsection{Hyperparameters} As discussed in \cite{NRDE}, there are two important hyperparameters that will heavily affect the performance and computational costs: the depth $N$ of the truncated log-signature and the number of steps (denoted by $n$ in \eqref{eqn:g_nrde}) for numerically solving the underlying ODE in \eqref{eqn:NRDE}.
To solve high-dimensional PPDEs, we intend to choose the optimal embedding to reduce the dimensionality of $X$ while preserving as much information about the original path as possible. Thus the reduced dimension $d_1$ of path $\mathcal{X}$ will also play a role and be treated as another important hyperparameter we need to tune.
About the network architecture of $\hat{f}$ that characterises the vector field of ODE in \eqref{eqn:NRDE}, we use grid search to find the optimal combination of network parameters that gives the best metric on test sets, as we discussed in next section.

\subsubsection{Packages} The package we use to calculate the log signature is the signatory \cite{signatory} We use the package provided by \cite{NRDE} in their paper to call and specify the hyperparameters of the NRDE network, and the numerical solver is provided by torchdiffeq \cite{torchdiffeq} wrapped in the NRDE module.

\subsubsection{Metrics on test sets}\label{sec:metrics}
We use two metrics to measure the model performance on test sets that contain simulated paths. The notation $u^{\text{truth}}(t,X)$ denotes the true solution of PPDE or the Monte-Carlo approximation (with $2000$ many simulations) based on the Feymann-Kac formula in \eqref{eqn:feynman} if the analytical form is not available. In accordance with the preceding sections, $\hat{u}_{\theta}(t,X)$ denotes the model's approximate solution to the PPDE and $\Pi=(t_j)_{j=0}^{N_1}$ the time discretisation of interval $[0,T]$. For simplicity, we set a fixed time stepsize $\Delta_{t}=t_{j+1}-t_{j}$. 
The absolute error on test sets with sampling size $N^{\text{test}}$ is defined as 
\begin{equation}
    \text{Abs.err}:=\frac{\Delta_{t}}{N^{\text{test}}}\sum_{i=1}^{N^{\text{test}}}\sum_{j=0}^{N_1}\left|u(t_j,X^{(i)}_{[0,t_j]})-\hat{u}_{\theta}(t_j,X^{(i)}_{[0,t_j]})\right|,
\end{equation}
and the relative error is 
\begin{equation}
    \text{Rel.err}:=\frac{\text{Abs.err}}{\frac{\Delta_t}{N^{\text{test}}}\sum_{i=1}^{N^{\text{test}}}\sum_{j=0}^{N_1}|u(t_j,X^i_{[0,t_j]})|}.
\end{equation}
Note that 
\begin{equation}
    \text{Abs.err}\approx \mathbb{E}\Big[\int_0^T \big|u(t,X_{[0,t]})-\hat{u}_{\theta}(t,X_{[0,t]})\big| \,\mathrm{d} t\Big]
\end{equation}
and
\begin{equation}
     \text{Rel.err}\approx \frac{\mathbb{E}\Big[\int_0^T \big|u(t,X_{[0,t]})-\hat{u}_{\theta}(t,X_{[0,t]})\big| \,\mathrm{d} t\Big]}{\mathbb{E}\Big[\int_0^T \big|u(t,X_{[0,t]})\big| \,\mathrm{d} t\Big]}. 
\end{equation}
The absolute error quantifies the total difference between the model output and the solution of PPDE on the whole path trajectory, and the relative error measures the scale of the solution hence we can compare the error across different dimensions. We choose $N^{\text{test}}=50$ and generate $10$ batches of such test sets so that we can calculate and compare the mean and variance of the two metrics across the different test sets.
\subsection{Heat Equation}\label{sec:heat}
The first experiment considered is a path-dependent heat equation with the following form 
\begin{equation}\label{eqn:bmppde}
       \left\{\begin{array}{l}\partial_t u\!\left(t,X_{[0,t]}\right)+\frac{1}{2}\operatorname{tr} \left[D_{x x} u(t,X_{[0,t]})\right]=0 \\ u\!\left(T,X_{[0,T]}\right)=g\!\left(X_{[0,T]}\right)\end{array}\right. ,
\end{equation}
Following Theorem \ref{thm:feyman}, the solution of the PPDE above is the expectation of the path-dependent final condition where $X$ is a $d$-dimensional Brownian motion. Given the following strong path-dependent terminal condition used in \cite{PDGM},
\begin{equation}
    g\!\left(X_{[0,T]}\right)=\left(\int_0^T \sum_{i=1}^d X^i(u) \, \mathrm{d} u\right)^2 ,
\end{equation}
where $X^i$ represent the $i$th coordinate of $X$, then PPDE \eqref{eqn:bmppde} has the analytical solution 
\begin{align*}
u\left(t,X_{[0,t]}\right)=&\left(\int_0^t \sum_{i=1}^d X^i(u) \,\mathrm{d} u\right)^2+2(T-t)\Big(\sum_{i=1}^d X^i(t)\Big) \int_0^t \sum_{i=1}^d X^i(u) \,\mathrm{d} u\\
    &+(T-t)^2\Big(\sum_{i=1}^d X^i(t)\Big)^2+\frac{d}{3}(T-t)^3.
\end{align*}
\begin{figure}\label{fig:evo}
    \centering
    \includegraphics[scale=0.35]{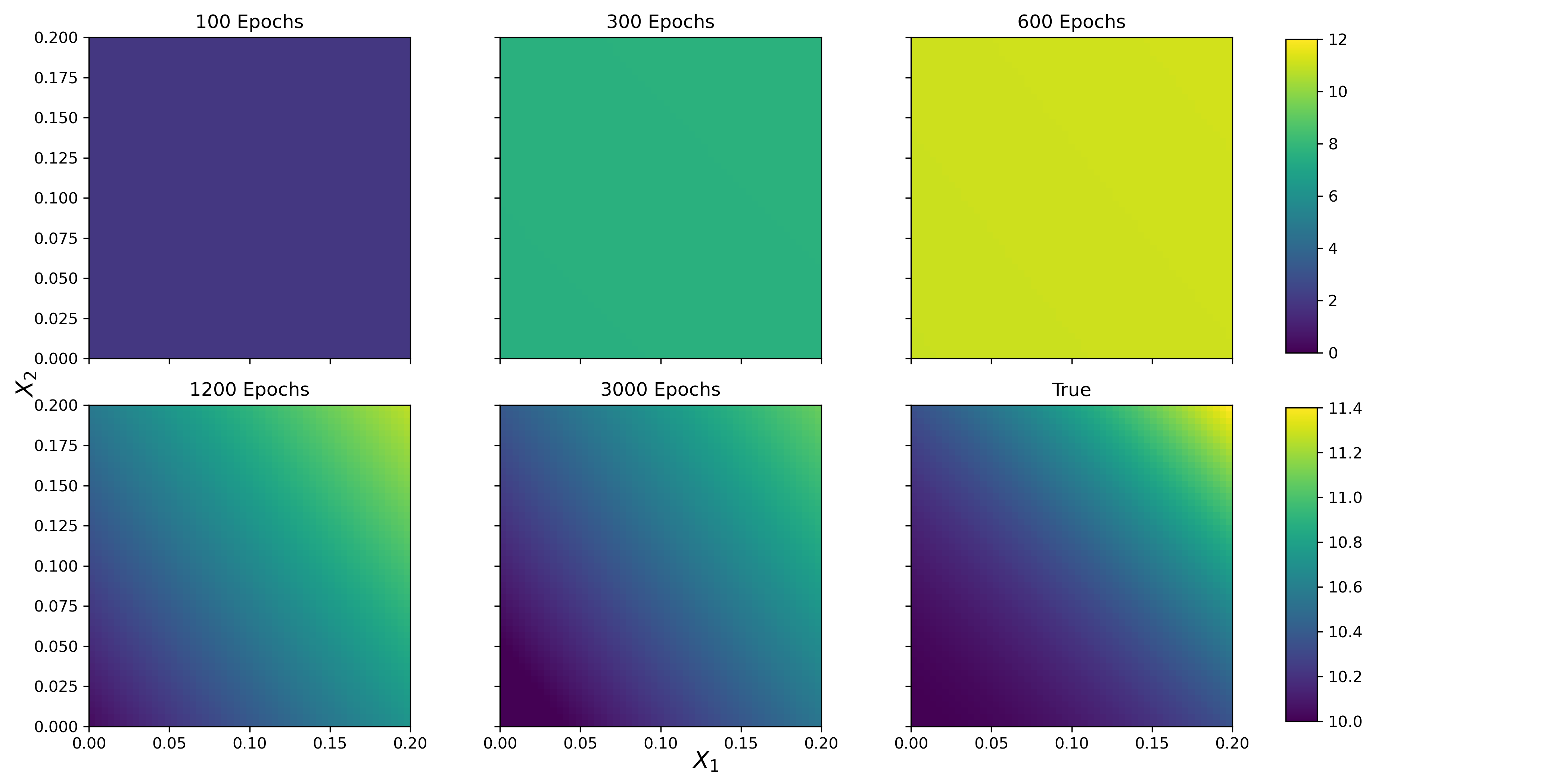}
    \caption{The evolution of model output $\hat{F}(t,X(t))$ to Eqn. \eqref{eqn:bmppde} at $t=0$, with the first two dimensions of $X(0) \in [0,0.2]^2$ and other dimensions=0. }
    \label{fig:epochs}
\end{figure}
Now fix the time interval $[0,1]$ and define $\Pi$ as a partition on the interval with step size $\Delta=0.1$ on which we will approximate the PPDE solution. We use a finer grid with time stepsize $\Delta_{\text{finer}}=0.01$ to simulate the underlying $X$.
    
We compare the performance of EL-SigLSTM and EL-NRDE models at $d=32$ under the learning task \emph{Method 2}. The network parameters are documented in Appendix \ref{sec:appc}.

\begin{figure}\label{fig:erro_32d}
    \centering
    \includegraphics[scale=0.25]{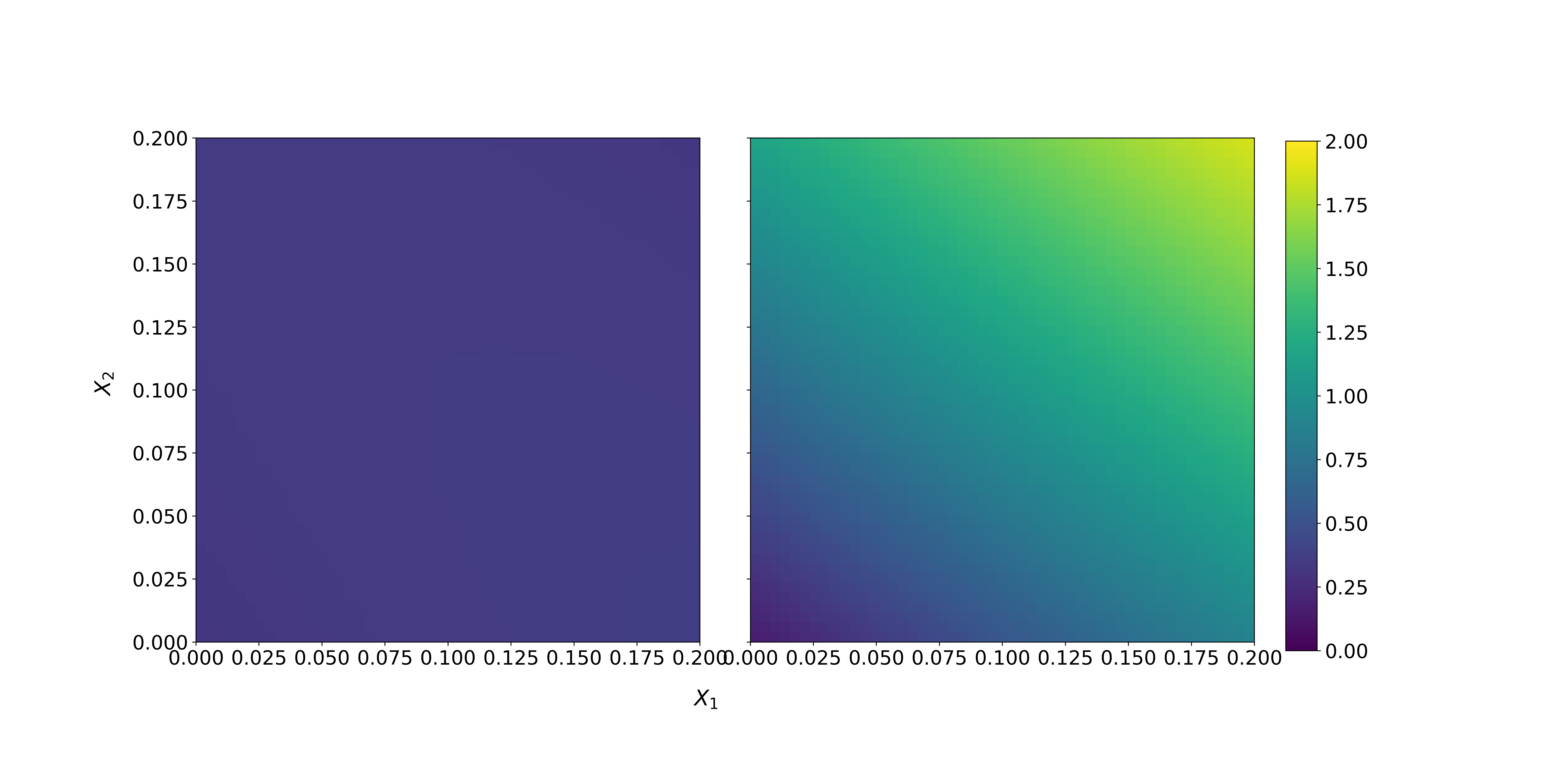}

    \caption{Error plots of EL-NRDE (left) and EL-LSTM (right) models for solving Eqn. \eqref{eqn:bmppde} on the whole spacial domain when $t=0$ and $d=32$. }
    \label{fig:heatmap_2d}
\end{figure}

To visualize the performance, we first explore the solution of PPDE at $t=0$, where we vary the first two dimensions of $X(0)$ and set the remaining dimensions to zero. Fig. \ref{fig:epochs} illustrates the evolution of the EL-NRDE's output at $t=0$ during the training process. The EL-NRDE network converges close to the true solution after 600 epochs, and the average relative approximation error on the whole spacial domain after training is $0.031$. This result is much better than the EL-SigLSTM model, as shown in Fig. \ref{fig:erro_32d}. 

\begin{figure}\label{fig:aggerr}
    \centering
    \begin{tabular}{cc}
        \includegraphics[scale=0.38]{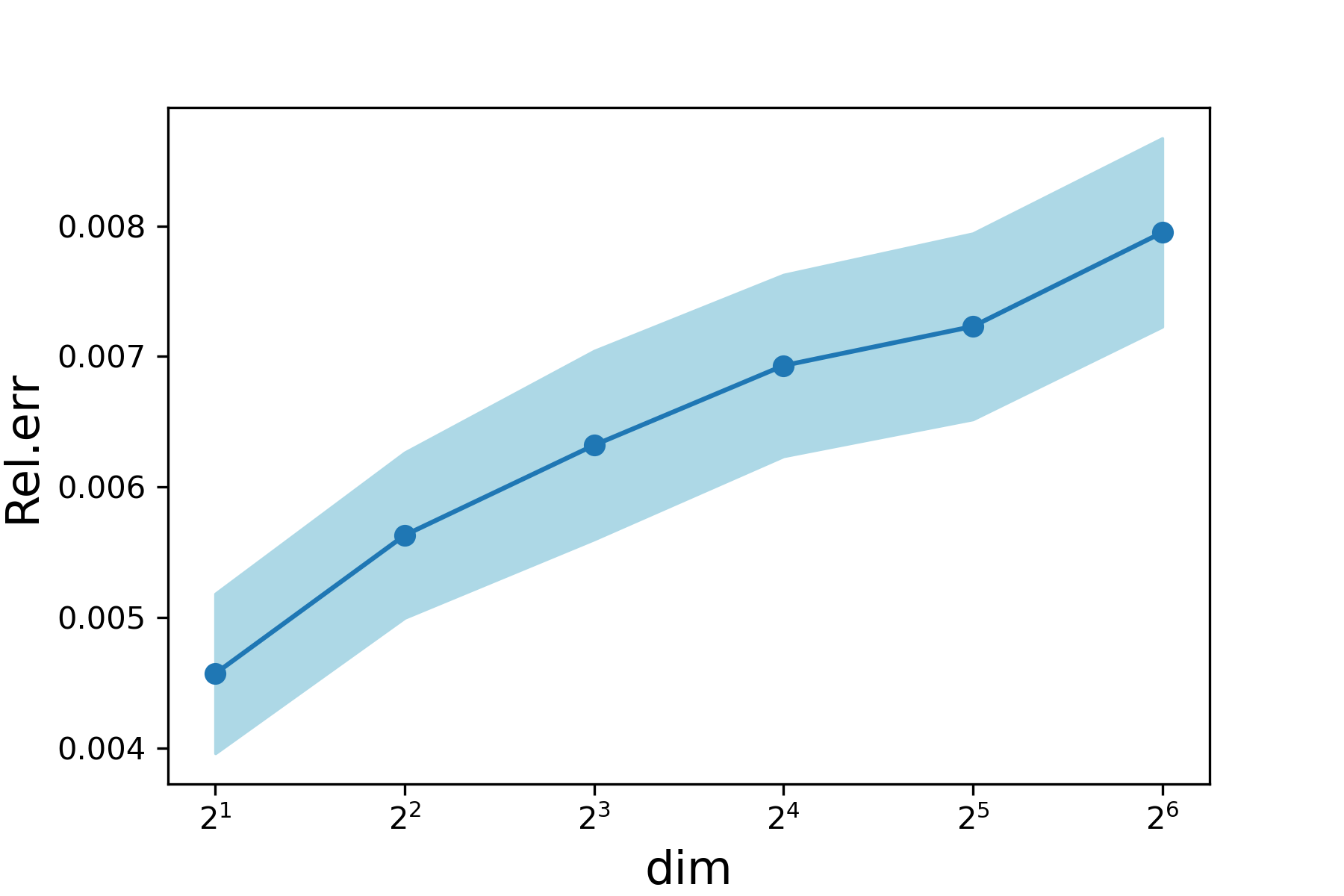} & \includegraphics[scale=0.38]{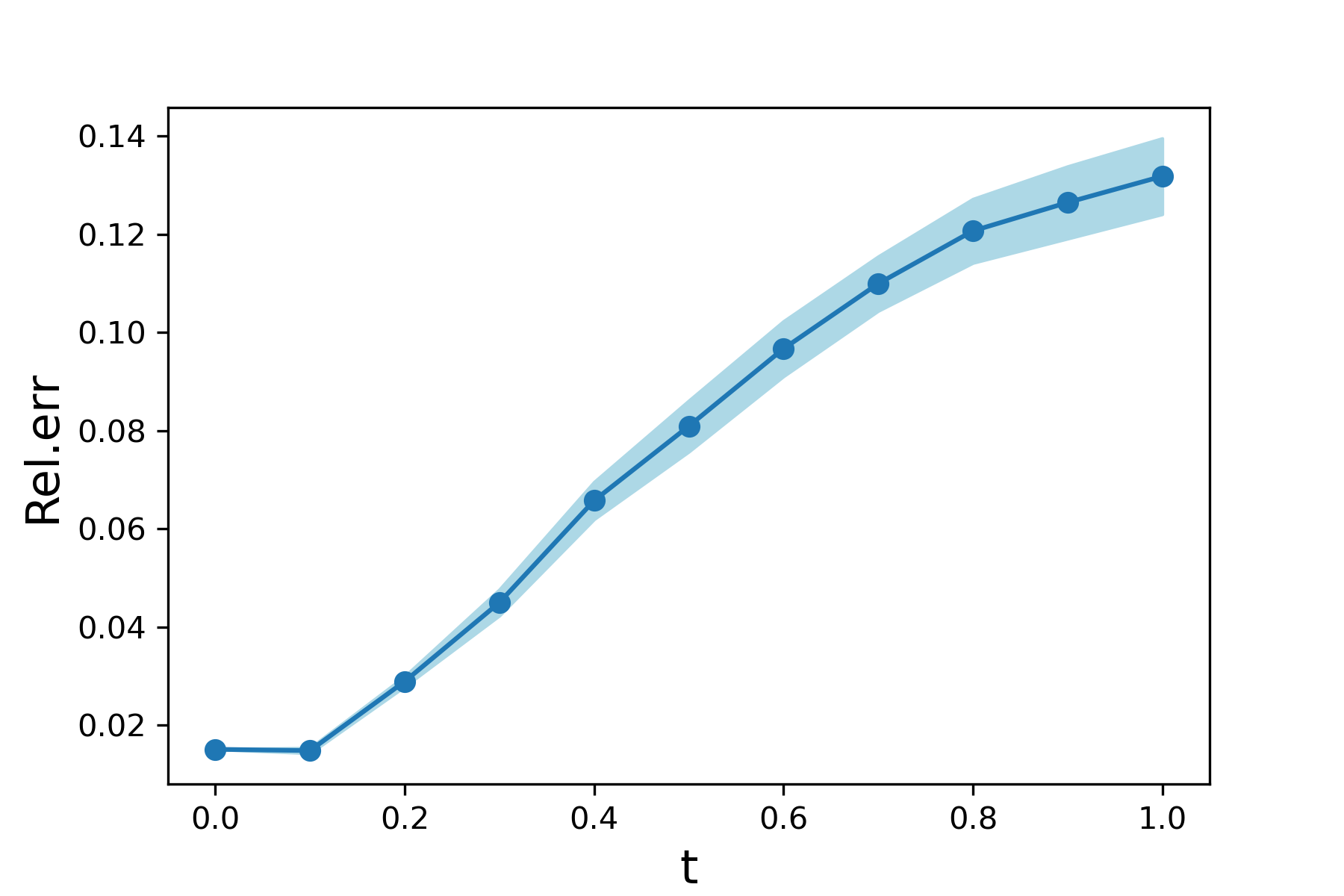} \\
    \end{tabular}
    \caption{Mean $\pm$ standard deviation of relative error from the EL-NRDE model for solving Eqn. \eqref{eqn:bmppde} among different dimensions (left) and at different times when $d=32$ (right).}
    \label{fig:my_label}
\end{figure}

We then compare the performance of two models across different dimensions. Table \ref{tab:heat_perf} compares the two test metrics introduced in Section \ref{sec:metrics} and the memory consumption during training. It is worth noting that the memory consumption for $d=8$ is comparable to that of $d=16$ for both models. This is due to the different optimal depths of the (log-)signature obtained for each $d$.
In terms of the accuracy of the approximation, the EL-NRDE model outperforms the EL-SigLSTM model, achieving smaller absolute error and relative error across all dimensionality considered. In particular, the performance is much enhanced in the higher-dimensional setting. For instance, at $d=64$, the errors from the EL-NRDE model is about one over fifty of the ones from the EL-SigLSTM model. The left figure in Fig. \ref{fig:aggerr}, which shows the mean and standard deviation of the relative error achieved among different dimensions by the EL-NRDE model, also suggests that our proposed model scales linearly with dimension. 

\begin{figure}[h!]\label{fig:path_eval}
    \centering
    \begin{tabular}{cc}
        \includegraphics[scale=0.38]{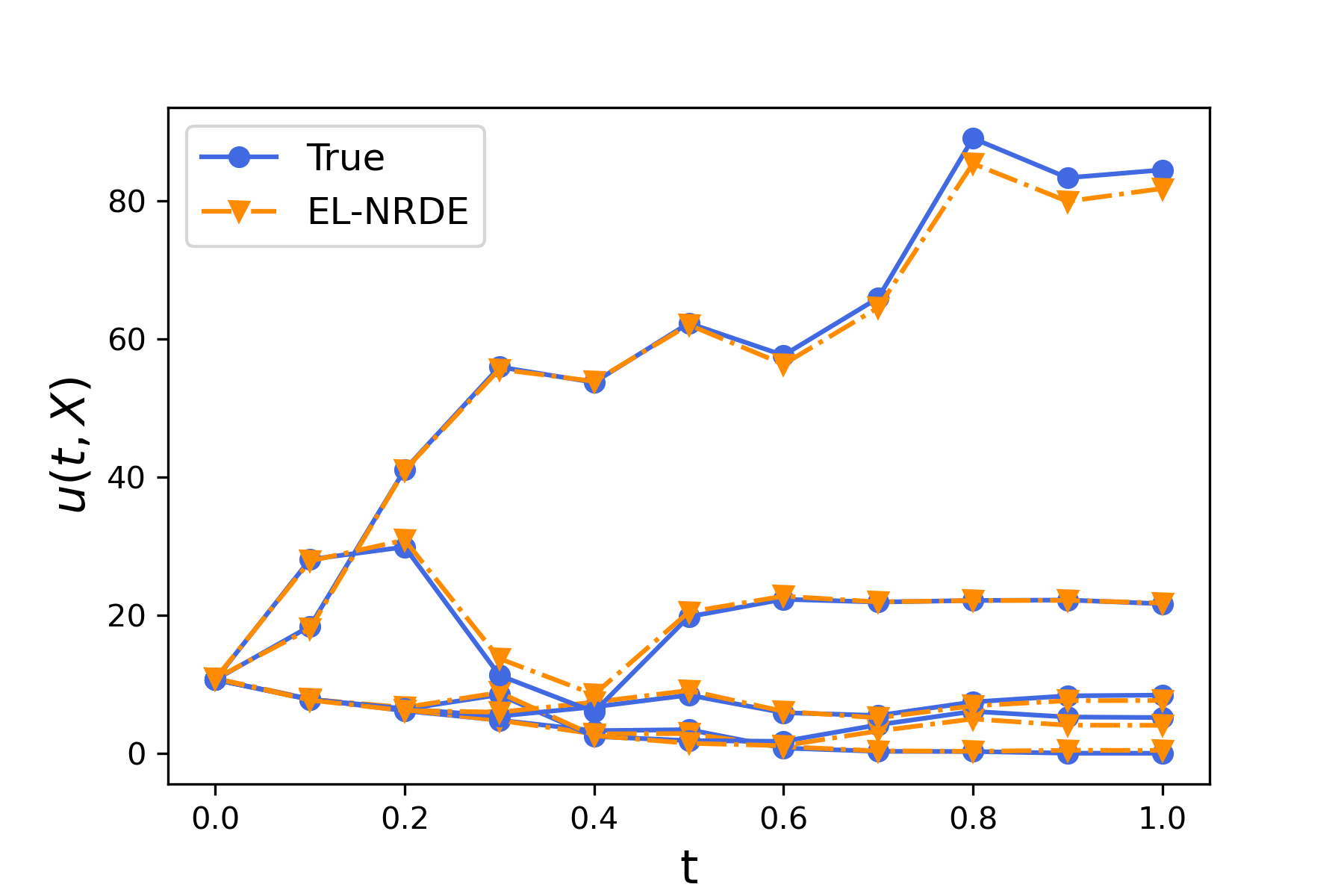} & \includegraphics[scale=0.38]{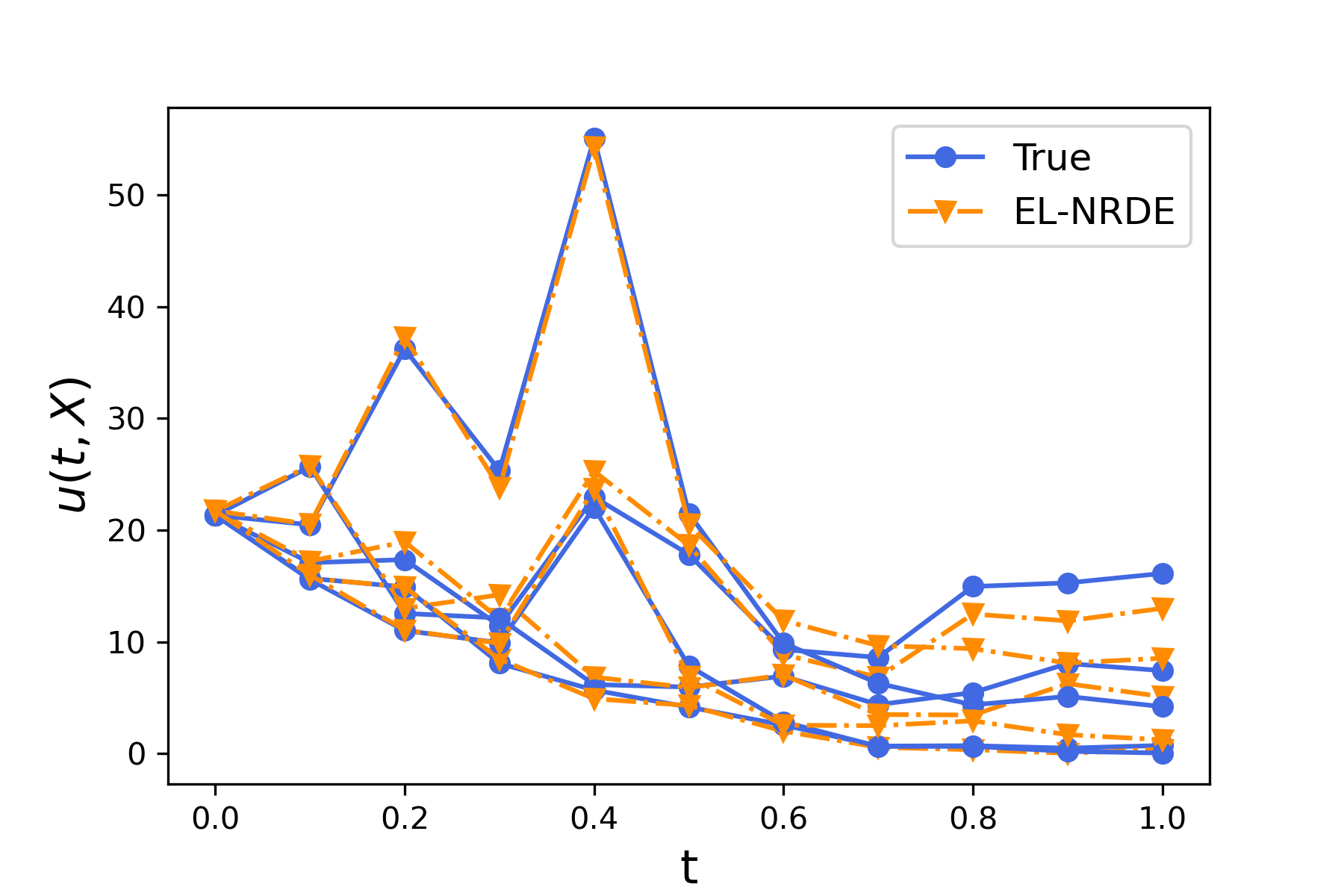} \\
    \end{tabular}
    \caption{True solutions of heat PPDE \eqref{eqn:bmppde} and their approximations generated by the \textbf{EL-NRDE model} on five different paths at $d=32$ (left) and $d=64$ (right). }
    \label{fig:Heat}
\end{figure}

We also demonstrate that the EL-NRDE model is capable of approximating $u(t,X)$ on the entire time interval by tracking the relative approximation error and its standard deviation on time partition $\Pi$, shown in the left figure in Fig. \ref{fig:aggerr}. In Fig. \ref{fig:path_eval}, we randomly simulate five trajectories of underlying path $X$ for $d=32$ and $d=64$ respectively and compare the model output to the exact solution. It can be observed that the approximations from the EL-NRDE model resemble the ground truth.
\begin{table}[h!]\label{tab:heat_perf}
    \center
    \begin{tabular}{@{}lcccccc@{}} \toprule
&\multicolumn{3}{c}{EL-SigLSTM}&\multicolumn{3}{c}{EL-NRDE} 
\\ \cmidrule(lr){2-4} \cmidrule(lr){5-7}
 Dim   &Abs.err& Rel.err& Memory  &Abs.err& Rel.err& Memory\\\midrule
8&0.2466 & 0.0078& 7.76&0.2096&0.0069& 5.21\\
16&1.523 & 0.0272 &  6.23 & 0.3170&0.0053&5.27 \\
32&2.45 & 0.0191 &  7.89 & 0.8427&0.0071&6.02 \\
64&15.66 & 0.0670 &  7.76 & 1.9610&0.0080&7.21\\

\bottomrule
\end{tabular}
    \caption{Performance of the EL-SigLSTM and EL-NRDE models on solving the heat PPDE \eqref{eqn:bmppde}.}
    \label{tab:bmppde}
\end{table}

\subsection{Black-Scholes Equation with Lookback option}\label{sec:lookback}
Following \cite{https://doi.org/10.48550/arxiv.2011.10630}, we consider the classical Black-Scholes model with a path-dependent payoff. Given a $d$-dimensional Brownian motion $W$, a risk-free rate $r$ and a symmetric positive-definite covariance matrix $\Sigma=LU$, where $LU$ are the lower-upper (LU) decomposition of $\Sigma$. The price process of risky assets $X\in \mathbb{R}^d$ in the market under the risk-neutral measure is the geometric Brownian motion with the following SDE
\begin{equation}\label{eqn:bseq}
    d X^i(t)=r X^i(t)\,\mathrm{d}t+\sigma^{i} X^i(t) \sum_{j} L^{i j} \,\mathrm{d} W^j(t),
\end{equation}
where $\sigma^i$ is the volatility of asset $i$. For the final payoff $u(T,X_{[0,T]})=g(X_{[0,T]})$, consider the lookback path-dependent payoff 
\begin{equation*}
    g(X_{[0,T]})=\max _{t \in[0, T]}\left[ \sum_{i} X^i(t)-\sum_{i} X^i(T)\right]\geq 0.
\end{equation*}
Define $\Sigma_{*}\in \mathbb{R}^{d\times d}$ where $\Sigma_{*}^{ij}=\sigma^i X^i L^{ij}$, then corresponding PPDE should be,
\begin{equation}\label{eqn:bsppde}
    \left\{\begin{array}{l}\partial_t u\left(t,X\right)+r(\partial_xu-u)+\frac{1}{2}\operatorname{tr}(D_{xx}u(t,X)\Sigma_*^T\Sigma_*)=0 \\ u\left(T,X_{[0,T]}\right)=g\left(X_{[0,T]}\right)\end{array}\right. .
\end{equation}
In the experiment, we take $r=5 \%, \Sigma^{i i}=1,T=1, \Sigma^{i j}=0,$ for $i \neq j$. Following \cite{https://doi.org/10.48550/arxiv.2011.10630}, the initial values $X(0)$ are sampled from a lognormal distribution 
\begin{equation*}
    X(0) \sim \exp \left(\left(\mu-0.5 \sigma^{2}\right) \tau+\sigma \sqrt{\tau} \xi\right),
\end{equation*}
where $\xi \sim \mathcal{N}(0,1), \mu=0.08, \tau=0.1$.

The setting for time discretization is the same as Section \ref{sec:heat}. The network parameters are documented in Appendix \ref{sec:appc}. We report performances from our model as well as the baselines for the low-dimensional case and high-dimensional case separately. 

\subsubsection{Low dimensional case} 
\begin{figure}
    \centering
    \begin{tabular}{cc}
        \includegraphics[scale=0.38]{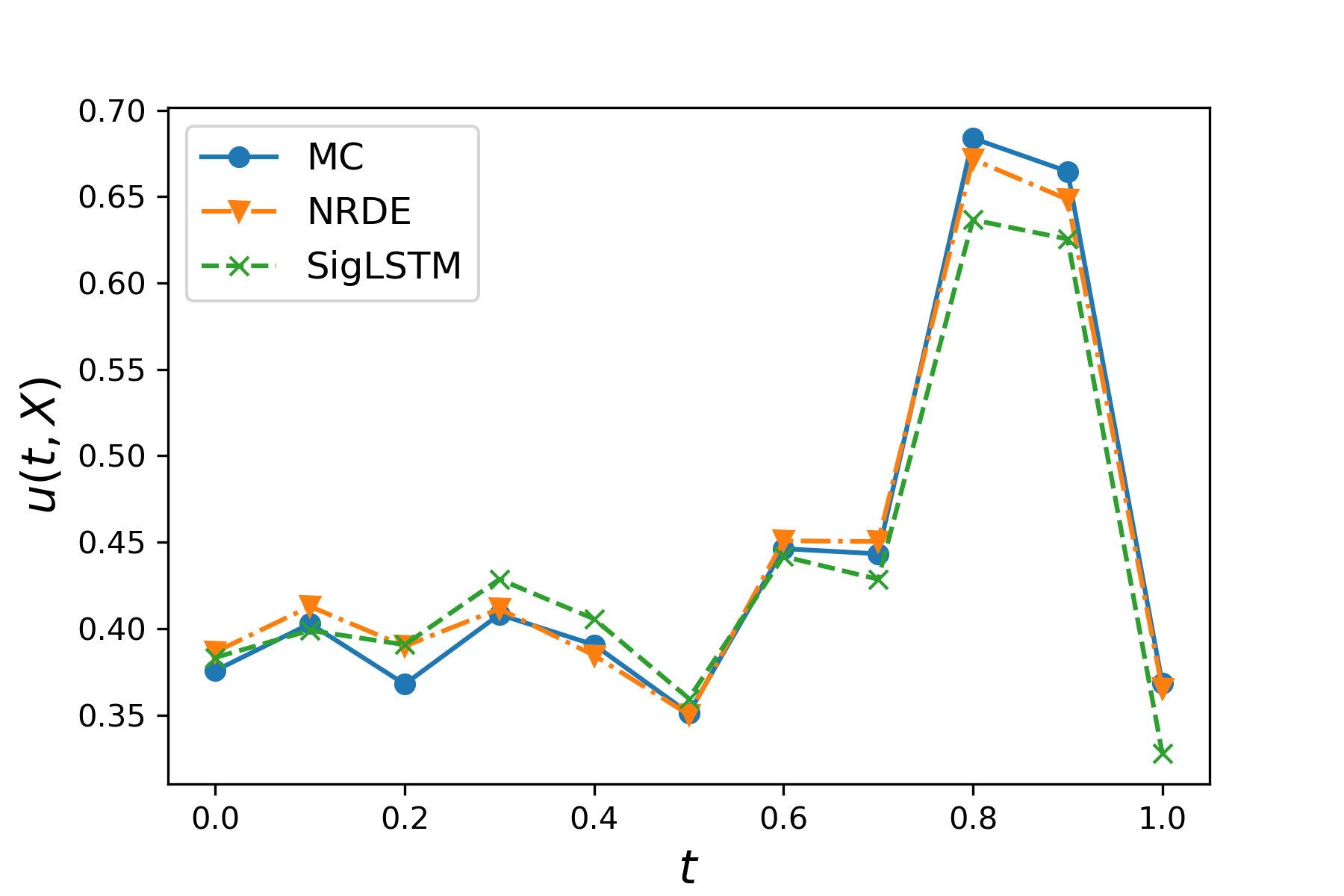} &\includegraphics[scale=0.38]{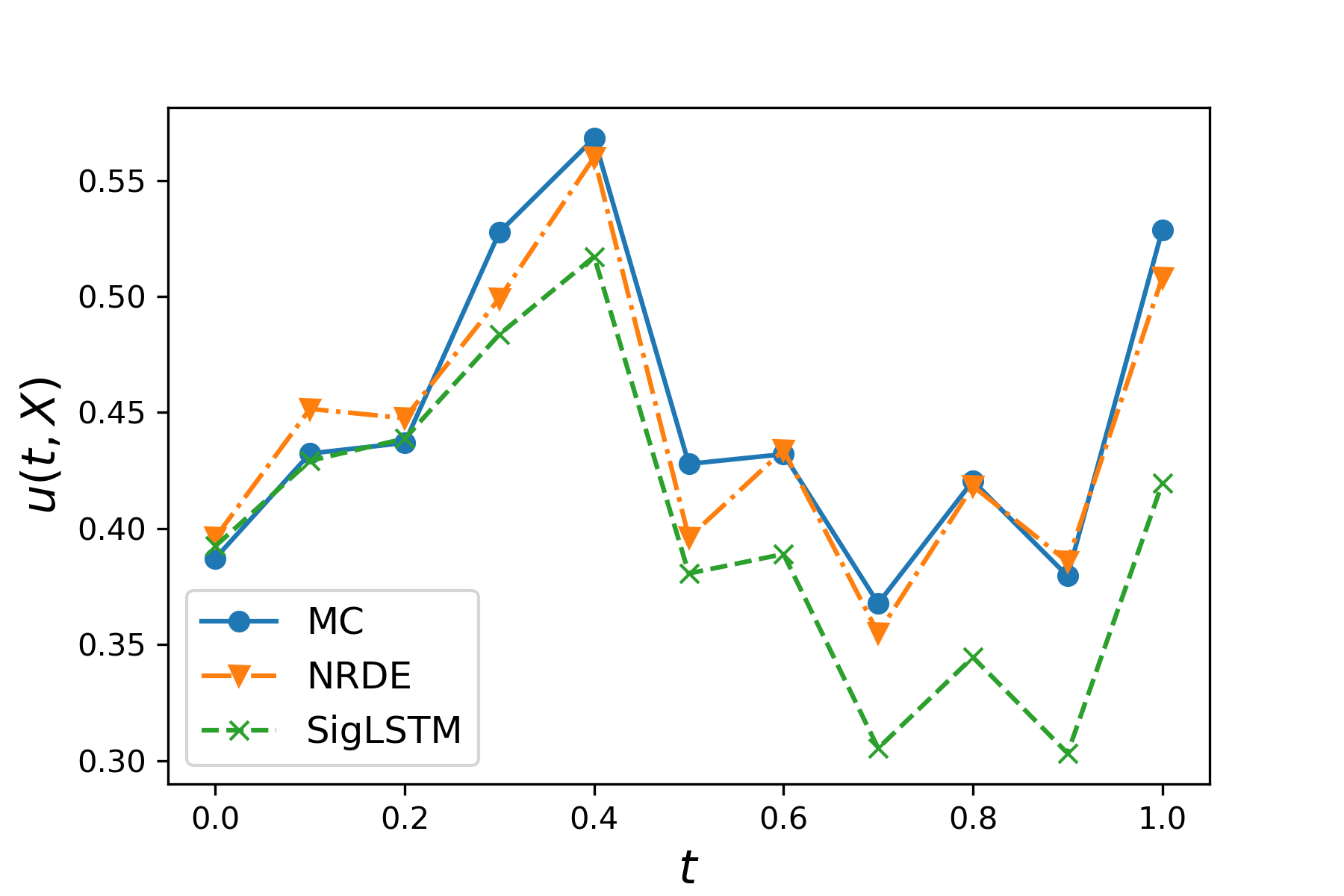}  \\
         
    \end{tabular}
    \caption{Comparison between outputs of \textbf{SigLSTM} and NRDE models on two different paths driven by \eqref{eqn:bmppde}, the "ground truth" here is given by the Monte-Carlo simulation.}
    \label{fig:BS_4d}
\end{figure}

We demonstrates the performance in the case of low-dimensionality of $X$ at $d=4$. The learning task adopted in this example is \emph{Method 1} in Section \ref{sec:method1} because we are using the default model without embedding, and we try to keep the whole network structure as simple as possible. We compare the prediction accuracy and memory efficiency between the baseline, say, the SigLSTM model, and our NRDE model.  As shown in table \ref{table:bs-4d}, the NRDE model outperforms the SigLSTM model in terms of accuracy, standard deviation and memory efficiency. In particular, the NRDE model reduces 30\% of the error from SigLSTM. Fig. \ref{fig:BS_4d} visualizes the model predictions of two models, where the reference solution is obtained by the Monte Carlo simulation.

\begin{table}
    \centering
    \begin{tabular}{@{}lccc@{}} \toprule
        &Abs.err ($\times 10^{-2}$) &Rel.err ($\times 10^{-3}$)&Memory (GIB) \\\midrule
        SigLSTM &$(3.65\pm0.18)$ &$(8.84\pm0.45)$&4.75\\
        NRDE & $(2.58\pm0.15)$ & $(5.95\pm0.31)$&3.63\\
        
    \bottomrule
    \end{tabular}
    \caption{Performance of SigLSTM model and NRDE model on Black-Scholes model driven by a four-dimensional Eqn. \eqref{eqn:bsppde}.}
\label{table:bs-4d}
\end{table}

\subsubsection{High dimensional case} 
\begin{figure}
    \centering
    \includegraphics[scale=0.5]{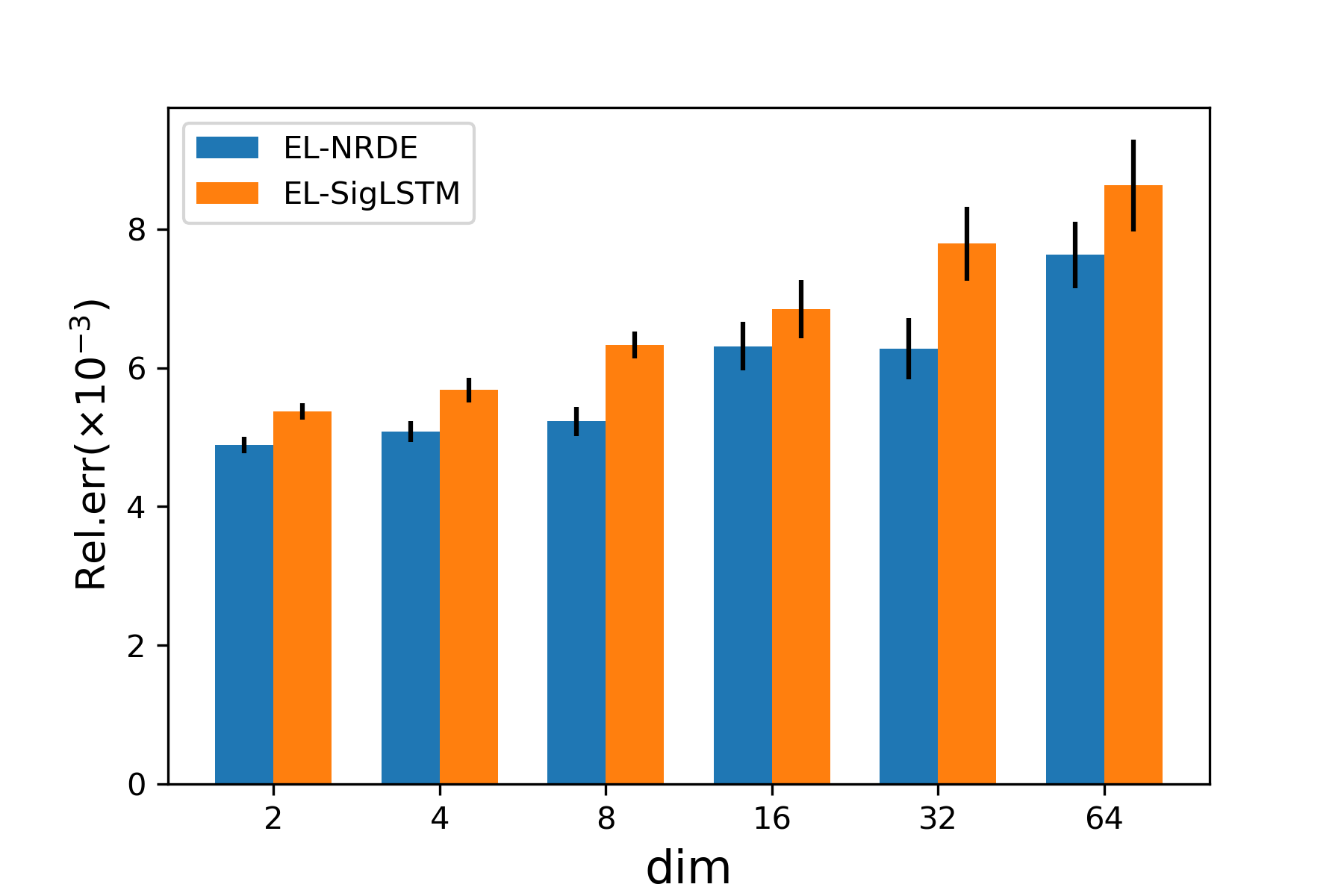}
    \caption{ Mean and standard variation of relative errors obtained by the EL-NDRE model and the EL-SigLSTM model among different dimensions for solving Eqn. \eqref{eqn:bsppde}.}
    \label{fig:dim_abserr}
\end{figure}

\begin{figure}[]
    \centering
    \begin{tabular}{cc}
        \includegraphics[scale=0.38]{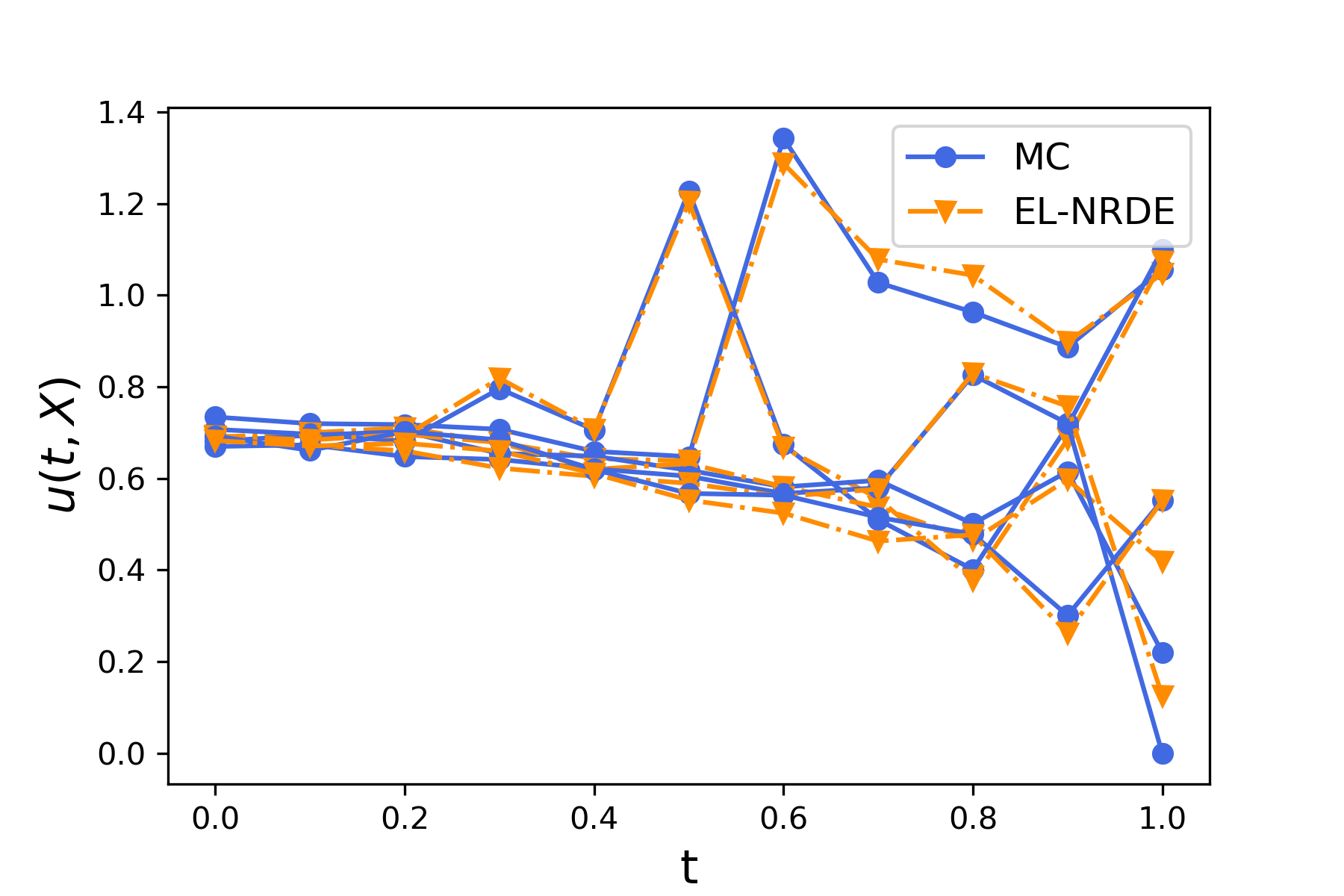} & \includegraphics[scale=0.38]{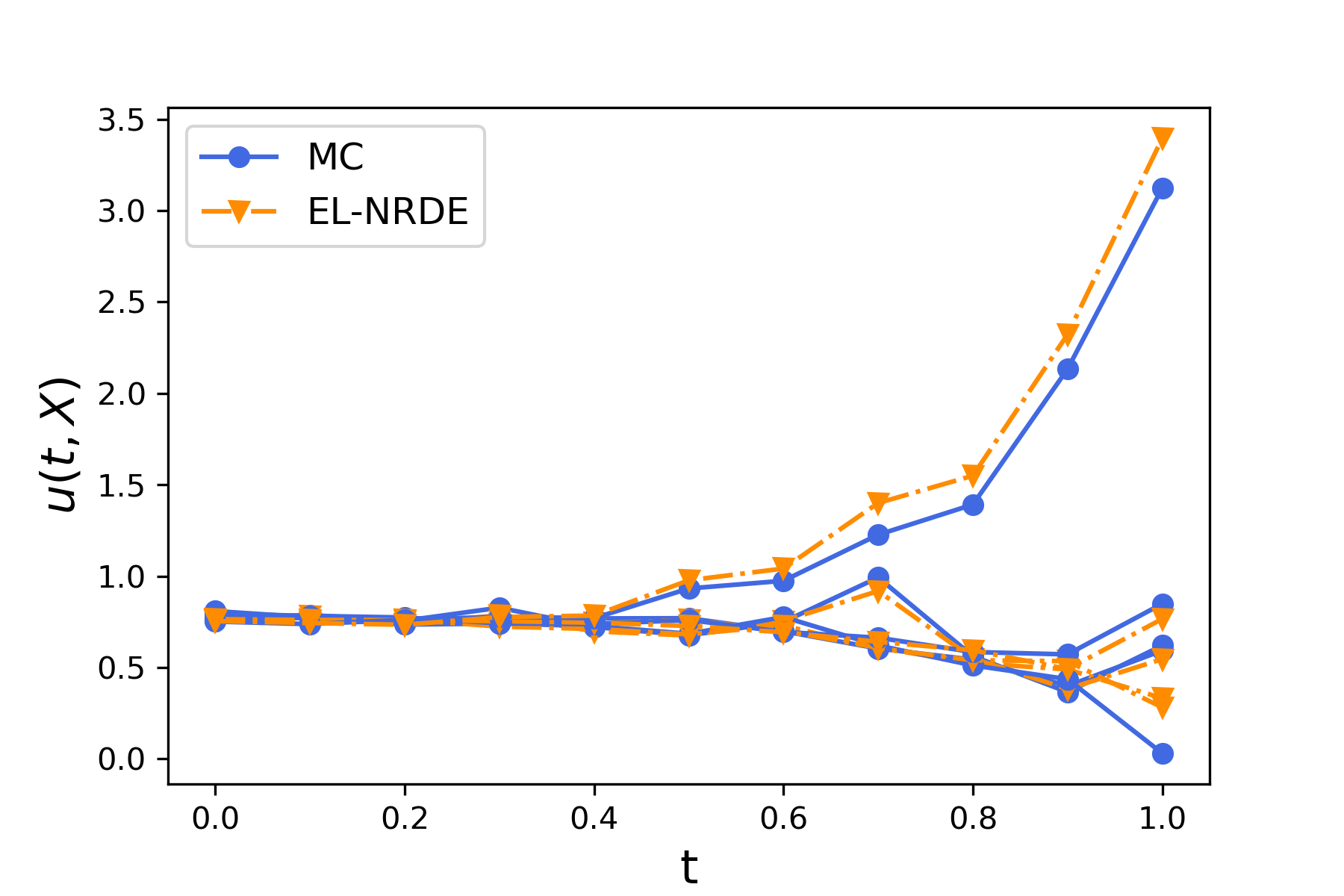}\\ 
    \end{tabular}
    \caption{Predictions of Black-Scholes PPDE solutions of \eqref{eqn:bsppde} on five different underlying paths given by \textbf{EL-NRDE} model and the Monte-Carlo simulation for $d=32$(left) and $d=64$ (right).}
    \label{fig:bs-hd}
\end{figure}

We now illustrate the scalability of our model with an embedding layer, say, the EL-NRDE model, under an increased dimensionality of the underlying process $X$. As shown in Table \ref{tab:BS-hd}, the dimension reduction feature is crucial when dealing with high dimensional PPDE: without the embedding, it is impossible to incorporate the signature feature via the SigLSTM model proposed in \cite{https://doi.org/10.48550/arxiv.2011.10630}. The EL-NRDE model reduces the approximation error by $10\%-15\%$ and it only consumes $63\%$ of memory used by EL-SigLSTM model during the training of $d=64$ experiment. Fig. \ref{fig:dim_abserr} compares the relative error obtained by the two models among different dimensions. Fig. \ref{fig:bs-hd} visualizes the ground truths given by Monte-Carlo simulations and the corresponding outputs of the EL-NRDE model.

\begin{table}[h!]
    \centering
    \begin{tabular}{@{}lccccccc@{}} \toprule
&\multicolumn{1}{c}{SigLSTM}&\multicolumn{3}{c}{EL-SigLSTM}&\multicolumn{3}{c}{EL-NRDE} 
\\\cmidrule(lr){2-2} \cmidrule(lr){3-5} \cmidrule(lr){6-8}
 Dim &  &Abs.err& Rel.err& Memory  &Abs.err& Rel.err& Memory\\\midrule
8& NA&0.0344& 0.0063 &  3.45 & 0.0296&0.0052&2.90\\
16& NA&0.0457&0.0068  &6.94& 0.0431&0.0063&3.51 \\
32&NA&0.0596 &0.0077  &  7.88 & 0.0487&0.0062&6.18 \\
64&NA&0.0823 & 0.0086&  9.32 & 0.0688&0.0076&5.91\\
\bottomrule
\end{tabular}
    \caption{Performance of LSTM and hybrid NRDE model on BS model \eqref{eqn:bsppde}.} \label{tab:BS-hd}
   
\end{table}

\newpage
\subsection{Heston Model with an autocallable option}
\begin{figure}
    \centering
    \begin{tabular}{cc}
        \includegraphics[scale=0.38]{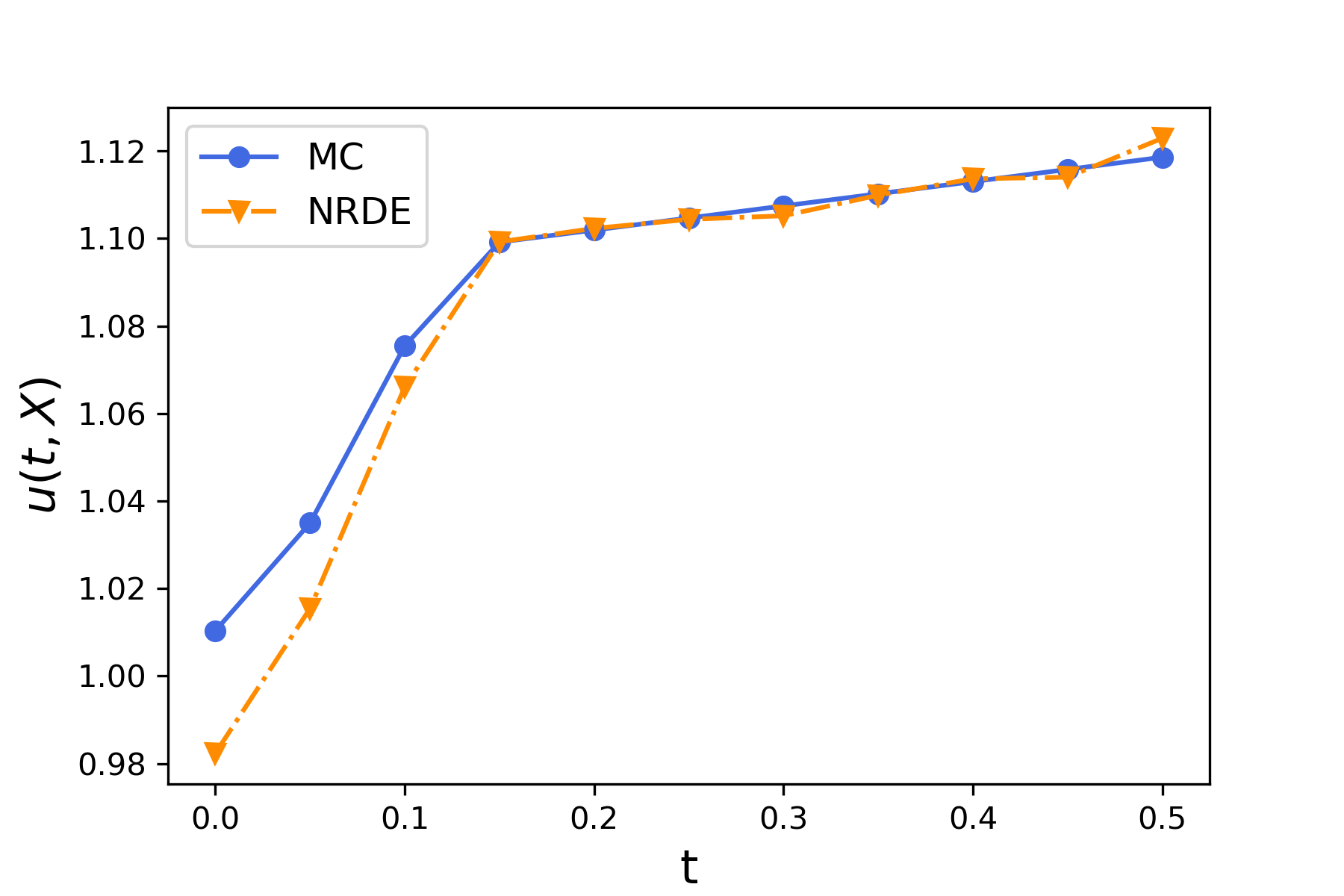} &\includegraphics[scale=0.38]{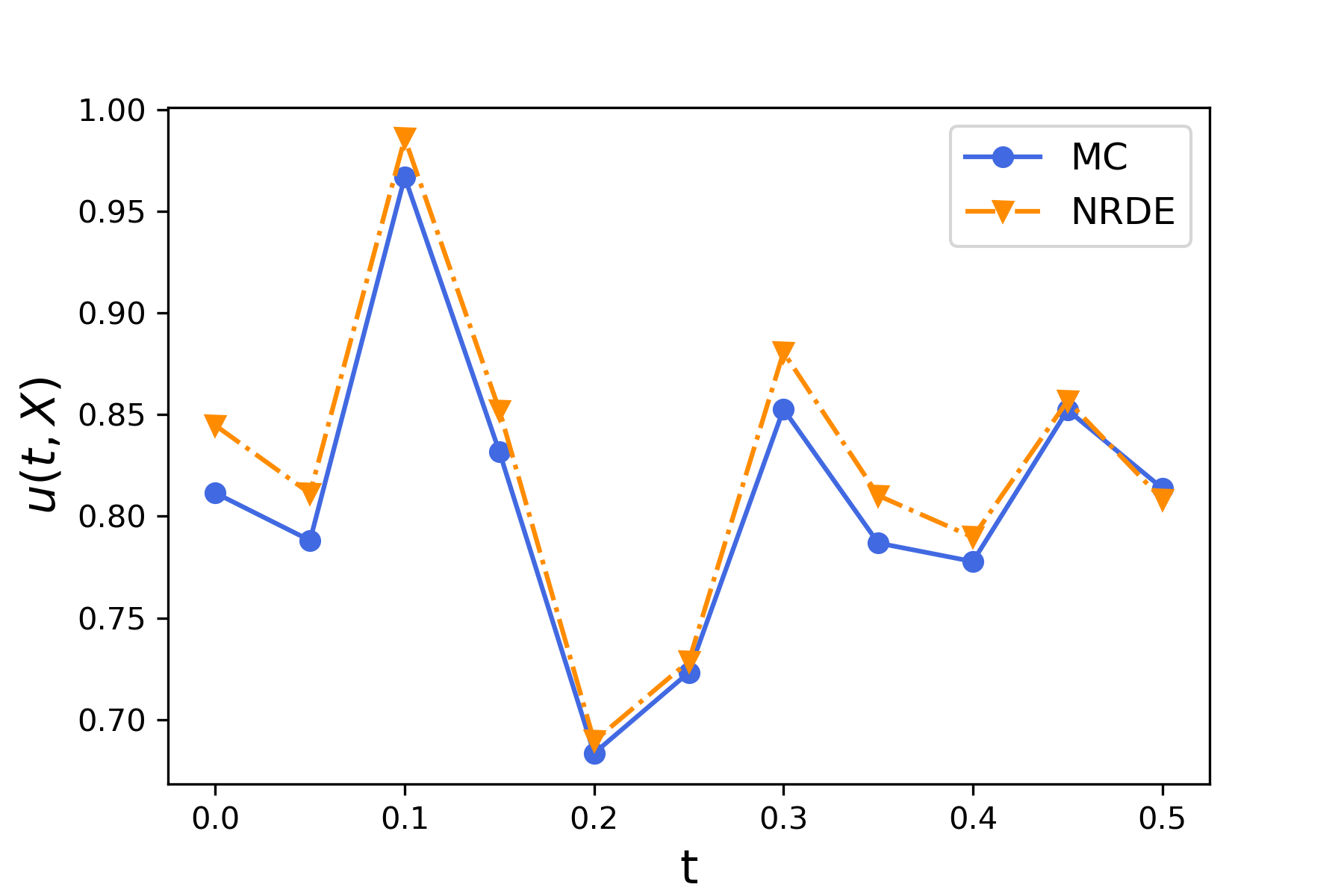}  \\
         
    \end{tabular}
    \caption{The illustration of the approximation of the NRDE model on one-dimensional Heston model \eqref{eqn:hstppde}, where the reference solution is obtained via the Monte-Carlo (MC) simulation.}
    \label{fig:Heston_2d}
\end{figure}

The last experiment, borrowed from \cite{https://doi.org/10.48550/arxiv.2011.10630}, is the Heston model with stochastic volatility, following the SDE
\begin{align}\label{eqn:heston}
\begin{split}
        &\mathrm{d}S(t)=\mu S(t)\,\mathrm{d}t+\sqrt{V(t)} S(t) \,\mathrm{d}W^s(t)\\
    &\mathrm{d}V(t)=\kappa(m-V(t))\,\mathrm{d}t+\eta\sqrt{V(t)}\,\mathrm{d} W^{v}(t),
\end{split}
\end{align}
where $S$ is the stock price, $V$ is the volatility and $W^s$ and $W^v$ are independent Brownian motions. We take $\mu=0.05$, $\kappa=0.8$, $m=0.3$ and $\eta=0.05$ and consider an autocallable payoff. Given the price information $S_{[0,T]}$,observation times $t_1\leq\dots\leq t_{N_1}$, a pre-defined barrier value $B$, premature payoff $Q_1,\dots,Q_{t_{N_1}}$ and a redemption payoff $q(\cdot)$, the discounted payoff of the univariate autocallable option is given by:
\begin{equation*}
    g\!\left(S_{[0,T]}\right)= \begin{cases}Q_j \quad \text { if } \quad S(t_i)<B \leq S(t_j) \quad \forall i<j \\ q\!\left(S(t_{t_{N_1}})\right) \quad \text { if } \quad S(t_j)<B \quad \forall j\end{cases}.
\end{equation*}
Now denote $X=(S,V)^T,W=(W^s,W^v)^T$, the Heston model \eqref{eqn:heston} can be rewritten in terms of $X(t)$: 
\begin{equation*}
   \mathrm{d} X(t)=\underbrace{\left(\begin{array}{c}
\mu S(t) \\
\kappa(m-V(t))
\end{array}\right)}_b \,\mathrm{d}t+\underbrace{\left[\begin{array}{c}
\sqrt{V(t)} S(t),  0 \\
0, \eta \sqrt{V(t)}
\end{array}\right]}_\sigma \,\mathrm{d} W(t),
\end{equation*}
with the corresponding PPDE
\begin{equation}\label{eqn:hstppde}
    \partial_tu(t,X)+b\partial_xu(t,X)+\frac{1}{2}\operatorname{tr}\left[D_{xx}u(t,X)\sigma^T\sigma\right]-\mu u(t,X)=0
\end{equation}
Following \cite{https://doi.org/10.48550/arxiv.2011.10630}, we consider a one-dimensional case only, where $S$ is the stock price of one asset, with a learning task as \emph{Method 1}. The parameters of the autocallable option is presented in Appendix \ref{sec:appc}. The table \ref{tab:hs_2d} shows that the NRDE model consistently outperforms the LSTM model, obtaining a relatively smaller approximation error and training loss, at the cost of about $25\%$ of memory. As Heston model is a more-involving model, both NRDE and SigLSTM may require more computational time to train compared with the linear SDE case in Section \ref{sec:lookback} as expected. Fig. \ref{fig:Heston_2d} illustrates the approximation result of the NRDE solver over the entire time interval driven by a simulated $(S,V)$. 

\begin{table}[]
    \centering
    \begin{tabular}{@{}lccc}\toprule
    &Abs.err ($\times 10^{-2}$)&Training loss ($\times 10^{-3}$)&Memory (GIB)\\\midrule
    NRDE&$1.45\pm 0.32$&$8.27$ &$0.524$\\
    SigLSTM&$1.78\pm 0.43$&$12.38$ &$2.25$\\
    \bottomrule
    \end{tabular}
    \caption{Performance of LSTM model and NRDE model on the Heston model \eqref{eqn:hstppde}.}
    \label{tab:hs_2d}
\end{table}

\newpage
\section{Conclusion} \label{sec:conc}
In this paper, we propose an efficient  neural-network-based model for solving high-dimensional PPDEs by encoding though a NRDE network the information of underlying random paths of PPDE and introducing an embedding layer for dimension reduction. Through various numerical examples, we present the significant performance of our model in accurately approximating the solution of the PPDE while demonstrating superior memory efficiency. In each experiment, our model consistently outperforms the baseline model, showcasing its remarkable capabilities.

In future research, it would be valuable to explore the theoretical properties of the NRDE solver in more depth. This could involve analyzing its consistency, speed of convergence, and stability. Furthermore, there is potential to enhance our approach by integrating advanced numerical schemes for simulating SDEs into our framework. Currently, we have relied on the Euler-Maruyama method for simulating underlying SDEs in the numerical section. However, for SDEs with irregular coefficients, this method may induce divergence issue.

\newpage

\newpage
\bibliographystyle{siamplain}

\newpage
\begin{appendix}\label{sec:appendix}
\section{The (log-)signature of a path}\label{sec:signature}
In this section, we expand the content in Section \ref{subsec:rough path} by defining the path signature in terms of the formal power series and providing the mathematical formula for a truncated log-signature. To help readers to understand the concept well, we will also give  an example of the signature and log-signature of a path. 

{\bf The signature in formal power series.} Let $\mathcal{I}=\{1,\dots,d\}$ be an index set of size $d$, $(i_1,\dots,i_k)$ with $ i_1,\dots,i_k\in \mathcal{I}$ be some multi-indexes, and $e_1,\dots,e_d$ be $d$ formal indeterminates. Then the algebra of non-commuting formal power series in $d$ indeterminates is the vector space of all series of the form 
\begin{equation}
    \sum_{k=0}^{\infty} \sum_{i_1, \ldots, i_k \in\{1, \ldots d\}} \lambda_{i_1, \ldots, i_k} e_{i_1} \ldots e_{i_k}, \quad \lambda_{i_1, \ldots, i_k} \in \mathbb{R}.
\end{equation}
Note that "non-commuting" means that $e_1e_2$ and $e_2e_1$ are distinct. The tensor algebra of $\mathbb{R}^d$ is the space of formal power series.
The addition and scalar multiplication in the vector space are naturally defined as 
\begin{align*}
    &\left(\sum_{k=0}^{\infty} \sum_{i_1, \ldots, i_k \in\{1, \ldots d\}} \lambda_{i_1, \ldots, i_k} e_{i_1} \ldots e_{i_k}\right)+\left(\sum_{k=0}^{\infty} \sum_{i_1, \ldots, i_k \in\{1, \ldots d\}} \mu_{i_1, \ldots, i_k} e_{i_1} \ldots e_{i_k}\right)\\
    &=\sum_{k=0}^{\infty} \sum_{i_1, \ldots, i_k \in\{1, \ldots d\}}\left(\lambda_{i_1, \ldots, i_k}+\mu_{i_1, \ldots, i_k}\right) e_{i_1} \ldots e_{i_k}
\end{align*}
and 
\begin{align*}
    a\left(\sum_{k=0}^{\infty} \sum_{i_1, \ldots, i_k \in\{1, \ldots d\}} \lambda_{i_1, \ldots, i_k} e_{i_1} \ldots e_{i_k}\right)=\sum_{k=0}^{\infty} \sum_{i_1, \ldots, i_k \in\{1, \ldots d\}} a \lambda_{i_1, \ldots, i_k} e_{i_1} \ldots e_{i_k}
\end{align*}
We can also define product between monomials $(e_{i_1}\dots,e_{i_k})$ and $(e_{j_1},\dots,e_{j_M})$
as 
\begin{equation}
    e_{i_1} \ldots e_{i_k} \otimes e_{j_1} \ldots e_{j_m}=e_{i_1} \ldots e_{i_k} e_{j_1} \ldots e_{j_m}.
\end{equation}
One can immediately see how the formal power series fits in with how we define the signature of the path $X:[0,T]:=J\rightarrow \mathbb{R}^d$. Again we have an index set with $d$ elements, and the monomials correspond to the order of our iterated integral 
\begin{equation}\label{eqn:iterated_integral}
S(X)_{J}^{i_1,\ldots,i_{k}}:=\int_{0<t_{k}<T}\cdots \int_{0<t_{1}<t_2} \text{d}X_{t_1}^{i_1}\ldots \text{d}X_{t_k}^{i_k},
\end{equation}
It is easy to verify the recursive relation
\begin{equation}
S(X)_{J}^{i_1,\ldots,i_{k}}=\int_{a<t_{k}<T} S(X)_{[0,t_k]}^{i_1,\ldots,i_{k-1}}\text{d}X_{t_{k}}^{i_{k}}
\end{equation}
and $S(X)_{J}^{i_1,\ldots,i_{k}}$ is an element of $X_J^k$ defined in \eqref{eqn:ite_int}.
 Then we have the formal power series representation of the signature as follows
\begin{equation}
    S_{J}(X)=\sum_{k=0}^{\infty} \sum_{i_1, \ldots, i_k \in\{1, \ldots d\}} S(X)_{J}^{i_1, \ldots, i_k} e_{i_1} \ldots e_{i_k}.
\end{equation}
{\bf Example A.1.} To demonstrate how to compute terms in $S_{J}(X)$, let us consider a two-dimensional path $X_t$ defined on interval $J:=[0,1]$ which can be parameterized by
\begin{align*}
    X(t)&=\left\{X^1(t), X^2(t)\right\}=\left\{t,t^2\right\}\\
    \mathrm{d}X(t)&=\left\{\mathrm{d}X^1(t), \mathrm{d}X^2(t)\right\}=\{\mathrm{d} t, 2t\, \mathrm{d} t\}.
\end{align*}
In this case the index set is $\mathcal{I}=\{1,2\}$ and the multi-index set is 
\begin{equation*}
W=\left\{\left(i_1, \ldots, i_k\right) \mid k \geq 1, i_1, \ldots, i_k \in\{1,2\}\right\}.
\end{equation*}
Then we can compute the terms in $S_J(X)$ simply as:
\begin{align*}
    S(X)_J^1 &=\int_{0<t<1}  \mathrm{d}  X^1(t)=\int_0^1 \mathrm{d} t=1,\\
    S(X)_{J}^2 &=\int_{0<t<1}  \mathrm{d} X^2(t)=\int_0^1 2t\, \mathrm{d} t=1,\\
    S(X)_J^{1,1} &=\int\int_{0<t_1<t_2<1}  \mathrm{d}  X^1(t_1) \,\mathrm{d}  X^1(t_2)=\int_0^1\left(\int_0^{t_2}  \mathrm{d}  t_1 \right)\, \mathrm{d} t_2=\frac{1}{2},\\
    S(X)_J^{1,2} &=\int\int_{0<t_1<t_2<1}  \mathrm{d}  X^1(t_1)\, \mathrm{d}  X^2(t_2)=\int_0^1 2t_2\left(\int_0^{t_2}  \mathrm{d}  t_1\right)  \mathrm{d} t_2=\frac{2}{3},\\
    S(X)_J^{2,1} &=\int\int_{0<t_1<t_2<1}  \mathrm{d}  X^2(t_1)\, \mathrm{d} X^1(t_2)=\int_0^1\left(\int_0^{t_2} 2t_1 \mathrm{d}  t_1\right) \mathrm{d} t_2=\frac{1}{3},\\
    S(X)_J^{2,2} &=\int\int_{0<t_1<t_2<1}  \mathrm{d}  X^2(t_1) \,\mathrm{d} X^2(t_2)=\int_0^1 2t_2\left(\int_0^{t_2} 2t_1 \mathrm{d} t_1\right)  \mathrm{d} t_2=\frac{1}{2},\\
    S(X)_J^{1,1,1} &=\int\int\int_{0<t_1<t_2<t_3<1}  \mathrm{d}  X^1(t_1) \,\mathrm{d}  X^1(t_2)\, \mathrm{d} X^1(t_3)=\frac{1}{6},\\
    &\dots
\end{align*}
Based on the calculation above, we have that $S_J^2(X)=(1,1,1,1/2,2/3,1/3,1/2)$.\\
Now if we plug in $S_J^2(X)$ into equation \eqref{eqn_tensor_logarithm} and truncated to degree of $2$, we get 
\begin{align*}
    &\log(S_J^2(X))\\
    &=0+e_1+e_2+\frac{1}{2}e_1 \otimes e_1+\frac{2}{3}e_1\otimes e_2+\frac{1}{3}e_2\otimes e_1+\frac{1}{2}e_2\otimes e_2\\
    &\quad -0-\frac{1}{2}(e_1+e_2)^{\otimes 2}\\
    &=0+e_1+e_2+0e_1 \otimes e_1+\frac{1}{6}e_1\otimes e_2-\frac{1}{6}e_2\otimes e_1+0e_2\otimes e_2\\
    &=e_1+e_2+\frac{1}{6}[e_1,e_2],
\end{align*}
where $[e_1,e_2]:=e_1\otimes e_2-e_2\otimes e_1$ and in the last line we change the basis of $T((E))$ to the basis of the subspace of $T((E))$ where log-signature lives\footnote{Indeed a log-signature takes value in the space of Lie series generated by $E$ \cite{lyons_differential_2007}.}, ie, 
$$\{e_1,e_2,[e_1,e_2],[e_1,[e_1,e_2]],[e_2,[e_1,e_2]],\ldots \}.$$
In summary, $\operatorname{Logsig}^2(X)=(0,1,1,1/6)$. Note that by elementary calculus, one can verify that 
$$S(X)^{i,i}_J=\frac{1}{2}(S(X)^i_J)^2 \text{ for }i\in \{1,2\}$$ and 
$$S(X)^{i,j}_J-\frac{1}{2}S(X)^i_JS(X)^j_J=\frac{1}{2}(S(X)^{i,j}_J-S(X)^{j,i}_J) \text{ for }i,j\in \{1,2\} \text{ and }i\neq j.  $$ 
Thus for the $2-$dimensional path $X$ we have that
\begin{equation*}
    \operatorname{Logsig}^2(X)=\Big(0,S(X)^1,S(X)^2,\frac{1}{2}(S(X)^{1,2}-S(X)^{2,1})\Big).
\end{equation*}

{\bf The dimension of truncated log-signature.} The length $\beta(d,N)$ of truncated log-signature $\operatorname{Logsig}^N(X)$ is given by $$\beta(d, N)=\sum_{k=1}^N \frac{1}{k} \sum_{i \mid k} \mu\left(\frac{k}{i}\right) d^i,$$ where $\mu$ is the Mobius function. Here $i|k$ means $i$ divides $k$ and the Mobius function is defined as 
\begin{equation*}
\mu(n)= \begin{cases}1, & \text { if } n=1; \\ (-1)^k, & \text { if } n \text{ is product of k primes};\\ 0; & \text { if } n \text{ has a sqaure factor greater than 1.}\end{cases}
\end{equation*}
For example, if $d=2$ and $N=2$, then $\beta(2,2)=3$, which is consistent with Example A.1.

\section{Adjoint state method for efficient gradient calculation}\label{sec:adjoint}
In this Section, we present the derivation of the adjoint state method that is used in the training of the NODE-type network via the Lagrangian multiplier approach \cite{plessix_review_2006,Koehler_Machine_Learning_and}.

The question is that, given our model $\hat{u}(t;\theta)$ and the loss functional $L(\hat{u};\theta)$, how to calculate the gradient $\frac{\mathrm{d} L}{\mathrm{d} \theta}$ efficiently. For the NODE network we have the following setup,
\begin{equation}\label{eqn:node_form}
    \frac{\mathrm{d} \hat{u}}{\mathrm{d} t}=\hat{f}(\hat{u}, t; \theta), \: \hat{u}(t=0)=u_0,
\end{equation}
where $\hat{u}(t;\theta)\in \mathbb{R}^N$ is the output of the model, $\hat{f}\in \mathbb{R}^N$ is the derivative characterized by some neural network and $\theta\in \mathbb{R}^P$ is the trainable parameter. Calculating $\hat{u}(t;\theta)$ is straightforward through a suitable numerical scheme like the Runge-Kutta or Euler method. If  define the loss function as $L(\hat{u},\theta)=\int_{0}^{T}g(u;\theta)\,\mathrm{d}t$ where $g\in \mathbb{R}$, then one can rewrite the gradient of the loss with respect to $\theta$ as
\begin{equation}\label{eqn:ode_loss}
    \frac{d L}{\mathrm{d} \theta}=\int_{0}^{T}\frac{d}{\mathrm{d} \theta} g(\hat{u} ; \theta) \,\mathrm{d} t=\int_{0}^{T}\Big(\frac{\partial g}{\partial \theta}+\frac{\partial g}{\partial \hat{u}}\frac{\mathrm{d} \hat{u}}{\mathrm{d} \theta} \Big)\mathrm{d}t,
\end{equation}
and our goal is to calculate $\frac{\mathrm{d}\hat{u}}{\mathrm{d} \theta}$ in \eqref{eqn:ode_loss} in a memory-efficient way. We first frame an optimization problem as follows
\begin{equation*}
    \underset{\theta}{\min} \ L(\hat{u};\theta) \;\;\;\;\; \text{ subject to } \: \hat{f}(\hat{u},t;\theta)-\frac{\mathrm{d} \hat{u}}{\mathrm{d}t}=0.
\end{equation*}
This is an equality-constrained optimization where we can use Lagrangian. Let $\lambda(t)\in \mathbb{R}^N$ be a continuous lagrangian multiplier, then we have 
\begin{equation*}
    \mathcal{L}:=L(\hat{u};\theta)+\int_{0}^{T} \lambda(t)^T(\hat{f}-\frac{\mathrm{d} \hat{u}}{dt})\mathrm{d}t=\int_{0}^{T} \Big(g(\hat{u};\theta)+ \lambda(t)^T(\hat{f}-\frac{\mathrm{d} \hat{u}}{\mathrm{d}t})\Big)\mathrm{d}t.
\end{equation*}
Since for any $ \theta\in \mathbb{R}^P, \hat{f}-\frac{\mathrm{d} \hat{u}}{\mathrm{d}t}=0$,  it holds that $\frac{\mathrm{d} \mathcal{L}}{\mathrm{d} \theta}=\frac{d L}{\mathrm{d} \theta}$. The objective now is to calculate
\begin{align*}
    \frac{\mathrm{d} \mathcal{L}}{\mathrm{d} \theta}&=\int_0^T \left(\frac{\partial g}{\partial \theta}+\frac{\partial g}{\partial \hat{u}}\frac{\mathrm{d} \hat{u}}{\mathrm{d} \theta}+ \lambda(t)^T \left(\frac{\partial \hat{f}}{\partial \theta}+\frac{\partial \hat{f}}{\partial \hat{u}}\frac{\mathrm{d} \hat{u}}{\mathrm{d} \theta}-\frac{\mathrm{d}}{\mathrm{d}t}\frac{\mathrm{d} \hat{u}}{\mathrm{d} \theta} \right)\right)\mathrm{d}t\\
    &=\int_0^T \left(\frac{\partial g}{\partial \theta}+\lambda(t)^T\frac{\partial \hat{f}}{\partial \theta}+\left(\frac{\partial g}{\partial \hat{u}}+\lambda(t)^T\frac{\partial \hat{f}}{\partial \hat{u}}-\lambda(t)^T\frac{\mathrm{d}}{\mathrm{d}t} \right)\frac{\mathrm{d} \hat{u}}{\mathrm{d} \theta}\right)\mathrm{d}t.
\end{align*}
As we have incorporated the constraint into the object, the value of $\lambda(t)$ can vary so the formula above can be further simplified. For example, one can choose appropriate $\lambda(t)$ such that the coefficient in front of term $\frac{d \hat{u}}{d \theta}$ becomes zero.
Note that by using integration by part,
\begin{equation*}
    \int_0^T -\lambda(t)^T\frac{\mathrm{d}}{\mathrm{d} t} \frac{\mathrm{d} \hat{u}}{\mathrm{d} \theta}\mathrm{d}t= \left[-\lambda(t)^T\frac{\mathrm{d} \hat{u}}{\mathrm{d} \theta} \right]_0^T+\int_0^T\left(\frac{\mathrm{d} \lambda(t)}{\mathrm{d} t}\right)^{\top} \frac{\mathrm{d} \hat{u}}{\mathrm{d} \theta} \mathrm{d} t.
\end{equation*}
Thus $\frac{\mathrm{d} \mathcal{L}}{\mathrm{d} \theta}$ can be simplied to
\begin{align*}
    \frac{\mathrm{d} \mathcal{L}}{\mathrm{d} \theta}=&\int_0^T \left(\frac{\partial g}{\partial \theta}+\lambda(t)^T\frac{\partial \hat{f}}{\partial \theta}+\left(\frac{\partial g}{\partial \hat{u}}+\lambda(t)^T\frac{\partial \hat{f}}{\partial \hat{u}}+\left(\frac{\mathrm{d} \lambda(t)}{\mathrm{d}t} \right)^T \right)\frac{\mathrm{d}\hat{u}}{\mathrm{d}\theta}\right)\mathrm{d}t \\&+ \lambda(0)^T\frac{\mathrm{d} \hat{u}}{\mathrm{d} \theta}(0)-\lambda(T)^T\frac{\mathrm{d} \hat{u}}{\mathrm{d}\theta}(T).
\end{align*}
To set the coefficient before $\frac{d \hat{u}}{d \theta}$ to zero
we have the following backward linear system of ODEs of $\lambda(t)$:
\begin{equation}\label{eqn:ode_back}
    \frac{\mathrm{d}\lambda}{\mathrm{d}t}=-\left(\frac{\partial \hat{f}}{\partial \hat{u}} \right)^T\lambda-\left(\frac{\partial g}{\partial \hat{u}} \right)^T, \;\;\; \text{with} \:\lambda(T)=\bold{0}.
\end{equation}
In summary, to calculate $\frac{\mathrm{d} \mathcal{L}}{\mathrm{d} \theta}$, one shall use an appropriate ODE solver to first solve the forward ODE \eqref{eqn:node_form} then the backward one \eqref{eqn:ode_back}. By doing so we are left with the gradient $$\frac{\mathrm{d} \mathcal{L}}{\mathrm{d} \theta}=\frac{\partial g}{\partial \theta}+\lambda(t)^T\frac{\partial \hat{f}}{\partial \theta} + \lambda(0)^T\frac{\mathrm{d} u_0}{\theta},$$
where $\hat{u}(0)=u_0$ as the initial condition. The gradient equation can be solved using numerical methods such as  quadrature. For values like $\frac{\partial \hat{f}}{\partial u},\frac{\partial g}{\partial u},\frac{\partial g}{\partial \theta},\frac{\partial \hat{f}}{\theta}$, we can find it analytically or using automatic differentiation.  Note that when we increase the number of trainable parameters $\theta$, the dimensionality of the two ODEs that need to be solved is unchanged, meaning that the computational cost scales constantly with $P$.\\
For our NRDE model, the only variation is that for hidden state $Z$ on some time interval $[t_0,t]$, if we have time partition $t_0=r_0<r_1<\dots<r_m=t$ , then
\begin{align}
   & Z(t)=Z(t_0)+\int_{t_0}^{t}\hat{h}(\theta,X,Z(s),s)\,\mathrm{d}s\label{eqn:Zt}\\
    &\hat{h}(\theta,X,Z(s),s)=\hat{f}(\theta,Z)\frac{\operatorname{LogSig}^{\text{depth}}_{r_i,r_{i+1}}(X)}{r_{i+1}-r_i}\label{eqn:htheta},
\end{align}
for $s\in [r_i,r_{i+1}]$. This model can be solved via the NODE. To get the desired output dimension for the NRDE model, one just need to add a final linear layer to $Z$. Thus it still enjoys the memory-efficient advantage described above.

\section{Numerical experiments}\label{sec:appc}
We include the full breakdown of our experiment results here. For the hyperparameter of an NRDE network, "h1" denotes the dimension of the hidden state $Z$ described in equation \eqref{eqn:Zt}, "\# layers" and "h2" the number of layers and size of hidden neurons in each layer of the feedforward network that characterize $\hat{f}(\theta,Z)$ in equation \eqref{eqn:htheta}, "depth" the depth of log-signature and "ODE solver" the type of ODE solver we use for solving ODE \eqref{eqn:Zt}. If an embedding layer is added to the NRDE network, ie, if we train an EL-NRDE model, d2 denotes the reduced dimension. In Table \ref{tab:autocall-conf}, we collect the parameters of the autocallable option we considered.

\begin{table}[h!]
    \centering
    \begin{tabular}{@{}lcccccccc@{}}\toprule
         d&d2&h1&\# layers&h2&depth&ODE solver&Rel.err&\# parameters\\\midrule
         8&2&15&6&30&4&Midpoint&0.0069&33548\\
         16&2&15&6&30&2&Midpoint&0.0053&9382\ \ \\
         32&4&15&6&30&2&Midpoint&0.0071&19066\\
         64&4&15&6&30&2&Midpoint&0.0081&20730\\
        \bottomrule
    \end{tabular}
    \caption{Network parameters used for EL-NRDE model in heat PPDE \eqref{eqn:bmppde}.}
    \label{tab:Heat-cofig}
\end{table}

\begin{table}[h!]
    \centering
    \begin{tabular}{@{}lcccccccc@{}}\toprule
         d&d2&h1&\# layers&h2&depth&ODE solver&Rel.err&\# parameters\\\midrule
         8&4&25&6&30&2&Midpoint&0.0054&27616\ \ \ \\
         16&4&25&6&45&4&Midpoint&0.0063&479734\ \ \\
         32&6&25&6&45&4&Midpoint&0.0062&1685174\\
         64&2&35&6&45&2&Midpoint&0.0076&117932\ \ \\
        \bottomrule
    \end{tabular}
    \caption{Network parameters used for EL-NRDE model in Black-Scholes PPDE \eqref{eqn:bsppde}.}
    \label{tab:BS-cofig}
\end{table}

\newpage
\begin{table}[h!]
    \centering
    \begin{tabular}{@{}ll@{}}\toprule
        Parameter& Value \\\midrule
         Terminal time& $T=0.5$\\
         Barrier &B=1.02\\
         Observation dates& $t_1=\frac{1}{6}$, $t_2=\frac{1}{3}$\\
         Premature payoffs& $Q_1=1.1$, $Q_2=1.2$\\
         Redemption payoff& $q(S(T))=0.9S(T)$\\
         \bottomrule
    \end{tabular}
    \caption{Parameters of the autocallable option.}
    \label{tab:autocall-conf}
\end{table}
\end{appendix}

\newpage
\end{document}


\maketitle

\section[Proof of Thm]{Proof of \cref{thm:feyman}}
\label{sec:proof}

\begin{proof}
Now define
\begin{align*}
           F(t, X_t)&:= \mathbb{E}\Big[g(X_T)e^{-\int_0^T r(s,X_s)\mathrm{d}s} +\int_0^T f\left(s,X_s\right)e^{-\int_0^s r(h,X_h)\mathrm{d}h}\mathrm{d}s \big|X_{t \wedge T}=\omega_{t \wedge T}\Big]\\
           &=e^{-\int_0^t r(s,X_s)\mathrm{d}s}\mathbb{E}\Big[g(X_T)e^{-\int_t^T r(s,X_s)\mathrm{d}s} \\
           &\qquad +\int_t^T f\left(s,X_s\right)e^{-\int_t^s r(h,X_h)\mathrm{d}h}\mathrm{d}s \big|X_{t \wedge T}=\omega_{t \wedge T}\Big]\\
           &\qquad+\mathbb{E}\Big[\int_0^t f\left(s,X_s\right)e^{-\int_0^s r(h,X_h)\mathrm{d}h}\mathrm{d}s \big|X_{t \wedge T}=\omega_{t \wedge T}\Big]\\
           &=e^{-\int_0^t r(s,X_s)\mathrm{d}s}u(t,X_t)+\int_0^t f\left(s,X_s\right)e^{-\int_0^s r(h,X_h)\mathrm{d}h}\mathrm{d}s,
\end{align*}
where the last line is from \eqref{eqn:feynman} and $X_t$ is known at current time $t$. Now applying the functional It\^o formula [REF] yields that
\begin{align*}
    \mathrm{d}F(t,X_t)&= \mathrm{d}\big(e^{-\int_0^t r(s,X_s)\mathrm{d}s}u(t,X_t)\big)+\mathrm{d}\big(\int_0^t f\left(s,X_s\right)e^{-\int_0^s r(h,X_h)\mathrm{d}h}\mathrm{d}s\big)\\
    &=e^{-\int_0^t r(s,X_s)\mathrm{d}s} \Big(\nabla_{\omega} u(t,X_t)\mathrm{d}X(t)+\big(\partial_{t} u(t,X_t)-r(t,X_t)u(t,X_t)\big)\mathrm{d} t\Big)\\
    &\quad +\frac{1}{2}e^{-\int_0^t r(s,X_s)\mathrm{d}s} \nabla_{\omega}^2 u(t,X_t)\mathrm{d} \langle X(t)\rangle +f(t,X_t)e^{-\int_0^t r(s,X_s)\mathrm{d}s} \mathrm{d}t\\
    &=e^{-\int_0^t r(s,X_s)\mathrm{d}s} \Big( \left[\partial_{t} u+b \nabla_{\omega} u+\frac{1}{2} \operatorname{tr}\left[\nabla_{\omega}^{2} u \sigma^{*} \sigma\right]-r u\right](t, X_t)+f(t,X_t)\Big)\mathrm{d}t \\
    &\quad +e^{-\int_0^t r(s,X_s)\mathrm{d}s} \sigma(t,X_t) dW(t).
\end{align*}
As $F(t,X_t)$ is a martingale, we can conclude with Theorem \ref{thm:feyman}.
\end{proof}

\section{Additional experimental results}
\Cref{tab:foo} shows additional
supporting evidence. 

\begin{table}[htbp]
{\footnotesize
  \caption{Example table}  \label{tab:foo}
\begin{center}
  \begin{tabular}{|c|c|c|} \hline
   Species & \bf Mean & \bf Std.~Dev. \\ \hline
    1 & 3.4 & 1.2 \\
    2 & 5.4 & 0.6 \\ \hline
  \end{tabular}
\end{center}
}
\end{table}

\bibliographystyle{siamplain}
\bibliography{references}